\documentclass{article}

\usepackage{lmodern}
\usepackage{arxiv}

\usepackage[utf8]{inputenc}
\usepackage[T1]{fontenc}
\usepackage{amsmath,amssymb,amsthm}
\usepackage{graphicx}
\usepackage[numbers,sort&compress]{natbib}
\usepackage{url}
\usepackage{microtype}
\usepackage{hyperref}
\hypersetup{colorlinks=true, linkcolor=blue!50!black, citecolor=blue!50!black, urlcolor=blue!50!black}

\newcommand{\Description}[1]{}

\theoremstyle{definition}
\newtheorem{definition}{Definition}

\usepackage{longtable}

\usepackage{enumitem}
\usepackage{comment}
\usepackage{siunitx}
\usepackage{forest}

\usepackage{colortbl}
\usepackage{xcolor}

\definecolor{specbg}{RGB}{238,237,254}   
\definecolor{verifbg}{RGB}{225,245,238}  
\definecolor{enfbg}{RGB}{250,236,231}    

\usepackage{multirow}              
\usepackage{booktabs}              
\usepackage{longtable}             
\usepackage{tabularx}
\usepackage{array}
\usepackage{makecell}
\usepackage{rotating}
\newcolumntype{L}[1]{>{\raggedright\arraybackslash}p{#1}}
\newcolumntype{C}[1]{>{\centering\arraybackslash}p{#1}}

\usepackage{tikz}
\usetikzlibrary{positioning, shapes.geometric, arrows.meta, calc, mindmap,shadows}

\definecolor{amber}{rgb}{0.75, 0.50, 0.00}
\definecolor{teal}{rgb}{0.00, 0.50, 0.50}
\definecolor{reviewerblue}{RGB}{0,70,140}
\definecolor{commentgray}{RGB}{240,240,245}
\definecolor{responsegreen}{RGB}{0,100,60}
\definecolor{grpFormal}{RGB}{180,160,220}   
\definecolor{grpPlan}  {RGB}{140,210,185}   
\definecolor{grpAgent} {RGB}{250,185,130}   
\definecolor{grpRun}   {RGB}{250,215,100}   
\definecolor{grpBench} {RGB}{195,195,185}   

\usepackage[acronym, nogroupskip, nonumberlist, nopostdot]{glossaries}
\glsdisablehyper

\newacronym{ai}{AI}{Artificial Intelligence}
\newacronym{bdd}{BDD}{Binary Decision Diagram}
\newacronym{bdi}{BDI}{Belief-Desire-Intention}
\newacronym{bmc}{BMC}{Bounded Model Checking}
\newacronym{ccs}{CCS}{Calculus of Communicating Systems}
\newacronym{cegar}{CEGAR}{Counterexample-guided Abstraction Refinement}
\newacronym{cegis}{CEGIS}{Counter-example Guided Inductive Synthesis}
\newacronym{cot}{CoT}{Chain-of-Thought}
\newacronym{csp}{CSP}{Communicating Sequential Processes}
\newacronym{ctl}{CTL}{Computation Tree Logic}
\newacronym{ctmc}{CTMC}{Continuous-Time Markov Chain}
\newacronym{dblp}{DBLP}{Digital Bibliography \& Library Project}
\newacronym[longplural={Deterministic Finite Automata}]{dfa}{DFA}{Deterministic Finite Automaton}
\newacronym{doi}{DOI}{Digital Object Identifier}
\newacronym{dsl}{DSL}{domain-specific language}
\newacronym{dtmc}{DTMC}{Discrete-time Markov Chain}
\newacronym{esbmc}{ESBMC}{Efficient SMT-Based Context-Bounded Model Checker}
\newacronym{fsm}{FSM}{Finite-State Machine}
\newacronym{grade}{GRADE}{Grading of Recommendations Assessment, Development and Evaluation}
\newacronym{hsd}{HSD}{Hierarchical Semantic Decomposition}
\newacronym{htn}{HTN}{Hierarchical Task Network}
\newacronym{ieee}{IEEE}{Institute of Electrical and Electronics Engineers}
\newacronym{llm}{LLM}{Large Language Model}
\newacronym{ltl}{LTL}{Linear Temporal Logic}
\newacronym[longplural={Markov Decision Processes}]{mdp}{MDP}{Markov Decision Process}
\newacronym{ml}{ML}{Machine Learning}
\newacronym{mtl}{MTL}{Metric Temporal Logic}
\newacronym{nl}{NL}{Natural Language}
\newacronym{nlp}{NLP}{Natural Language Processing}
\newacronym{nusmv}{NuSMV}{New Symbolic Model Verifier}
\newacronym{os}{OS}{operating system}
\newacronym{owasp}{OWASP}{Open Worldwide Application Security Project}
\newacronym{pac}{PAC}{Probably Approximately Correct}
\newacronym{pat}{PAT}{Process Analysis Toolkit}
\newacronym{pctl}{PCTL}{Probabilistic Computation Tree Logic}
\newacronym{pddl}{PDDL}{Planning Domain Definition Language}
\newacronym{plc}{PLC}{Programmable Logic Controller}
\newacronym{rl}{RL}{Reinforcement Learning}
\newacronym{rlaif}{RLAIF}{Reinforcement Learning from AI Feedback}
\newacronym{rlhf}{RLHF}{Reinforcement Learning from Human Feedback}
\newacronym{sat}{SAT}{Boolean Satisfiability}
\newacronym{slaif}{SLAIF}{Supervised Learning from AI Feedback}
\newacronym{smt}{SMT}{Satisfiability Modulo Theories}
\newacronym{spin}{SPIN}{Simple Promela Interpreter}
\newacronym{ssr}{SSR}{Safe Success Rate}
\newacronym{stl}{STL}{Signal Temporal Logic}
\newacronym{tl}{TL}{Temporal Logic}

\definecolor{orcidgreen}{HTML}{A6CE39}
\newcommand{\orcid}[1]{\href{https://orcid.org/#1}{\raisebox{-0.15em}{\tikz[baseline]{\fill[orcidgreen] (0,0.42em) circle (0.42em); \node[white, font=\sffamily\bfseries\fontsize{5.5}{5.5}\selectfont, inner sep=0pt] at (0,0.42em) {iD};}}}}

\title{Toward Safe LLM Agents: A Survey of Specification, Verification, and Enforcement}
\renewcommand{\shorttitle}{Toward Safe LLM Agents}

\author{%
  Pierre Dantas\,\orcid{0000-0001-6390-9340}\\
  The University of Manchester, UK\\
  Federal University of Amazonas, Brazil\\
  \texttt{pierre.dantas@manchester.ac.uk}
  \And
  Lucas Cordeiro\,\orcid{0000-0002-6235-4272}\\
  The University of Manchester, UK\\
  Federal University of Amazonas, Brazil\\
  \texttt{lucas.cordeiro@manchester.ac.uk}
  \AND
  Ehsan Nowroozi\,\orcid{0000-0002-5714-8378}\\
  University of Greenwich, UK\\
  \texttt{ehsan.nowroozi65@ieee.org}
  \And
  Norbert Tihanyi\,\orcid{0000-0002-9002-5935}\\
  E\"{o}tv\"{o}s Lor\'{a}nd University, Hungary\\
  \texttt{ntihanyi@inf.elte.hu}
}
\date{}

\begin{document}

\maketitle

\begin{abstract}
    \gls{llm} agents take irreversible real-world actions, yet no system gives formally grounded, task-level safety guarantees for their plans. In a PRISMA 2020 systematic review, we searched six databases (20--29 May 2026) for work on \gls{llm} agent specification, verification, and enforcement published in 2022--2026. We narratively synthesized 38 studies, assessed risk of bias per study, and graded evidence certainty with GRADE. Three authors independently screened a repeated search to validate the single-reviewer screening (agreement: Fleiss' $\kappa = 0.81$). First, specification is the bottleneck: on one benchmark, only \SIrange{24.8}{34.2}{\percent} of \gls{llm} translations from natural language into \gls{pddl} are semantically correct. Second, runtime monitoring is the most mature enforcement strategy; it reduced unsafe actions by \SI{65.37}{\percent} in one household-agent study, but its guarantees cover only single steps or short horizons. Third, in one $\tau$-Bench study of transactional tool-use agents, blocking \SI{94}{\percent} of unsafe actions still left fewer than \SI{5}{\percent} of tasks completed both safely and correctly in most settings, because agents found other unsafe routes (the verifier tax). Fourth, no approach simultaneously achieves soundness, completeness, scalability, semantic grounding, and utility preservation. For seven of the nine principal claims, GRADE gives the lowest certainty rating (Very Low), meaning further studies are likely to produce substantially different figures; each rests on one or two studies in narrow domains. Two of the open problems we rank highest are translating requirements into formal specifications that keep their meaning and enforcing safety over whole plans rather than single actions. We contribute a three-level taxonomy and an agenda of ten open problems.
\end{abstract}


\glsresetall


\section{Introduction}
\label{sec:intro}

\gls{llm} agents, systems in which an \gls{llm} acts as a reasoning engine that selects and sequences tool calls, sub-tasks, or robotic actions, have moved rapidly from research prototypes to production deployments. They now operate in consequential settings including software engineering~\citep{Liu2023agentbench}, healthcare~\citep{Miculicich2025veriguard}, customer service~\citep{Kamath2025agentc}, and autonomous driving~\citep{Wang2025probguard}. In each of these domains, an agent's plan, the sequence of actions it commits to executing, can have irreversible real-world effects: a misconfigured deployment, an erroneous medication recommendation, an unauthorized financial transaction, a collision.

The fundamental safety challenge is that \glspl{llm} generate plans through statistical pattern matching over training data, rather than through sound logical inference. A plan that reads fluently and appears coherent may violate safety invariants, disregard temporal ordering constraints (e.g., authentication must precede data access), or produce cascading harmful effects that are individually plausible yet collectively catastrophic. Benchmark evidence documents this gap: \emph{none} of 16 popular \gls{llm} agents scored above \SI{60}{\percent} on a structured safety benchmark spanning 2,000 test cases~\citep{Zhang2024agentsafety}. Frontier models show limited robustness even to basic jailbreak attacks in multi-step agentic settings~\citep{Andriushchenko2024agentharm}.

Classical \gls{ai} planning as well as formal verification offer mathematically rigorous tools for guaranteeing plan correctness: \gls{pddl}-based planners, model checkers such as \gls{spin} and \gls{nusmv}, and deductive verifiers such as Dafny and Coq. However, applying these tools to \gls{llm} agent outputs requires closing the gap between unstructured \gls{nl} and formal mathematical structures, a translation step that itself introduces semantic errors. The consequence is a \emph{specification-verification-enforcement pipeline} for \gls{llm} agent plans that is fragile at every stage: specifications may be semantically incorrect~\citep{Anon2025pddl_survey}, verification may operate over inaccurate formal models~\citep{Ramani2025bridging}, and enforcement may block unsafe single actions while leaving task-level safety unresolved~\citep{Anon2026verifiertax}.

This survey systematically maps the emerging literature that attempts to close this gap. We cover the period 2022 to 2026, encompassing the \gls{llm}-agent era from GPT-3/4 onward, and identify 38 studies across the specification, verification, enforcement, and evaluation subfields using a PRISMA 2020 systematic review protocol.

\subsection{Problem Statement}
\label{sec:intro:problem}

This survey is organized around four research questions:

\begin{description}
    \item[RQ1 (Specification):] What is the semantic correctness of natural-language-to-formal translation for LLM agent safety properties, and how does this bound downstream verification?
    \item[RQ2 (Verification):] Which verification paradigms (model checking, static analysis, theorem proving, LLM-assisted) are applied to LLM agent plans, and what are their scalability limits?
    \item[RQ3 (Enforcement):] Does blocking unsafe individual actions at runtime reliably produce safe task-level completion?
    \item[RQ4 (Evidence base):] What is the current strength of evidence (GRADE certainty) for the field's principal empirical claims, and what limits it?
\end{description}

\begin{figure}[htbp]
    \centering
    \includegraphics[width=\linewidth]{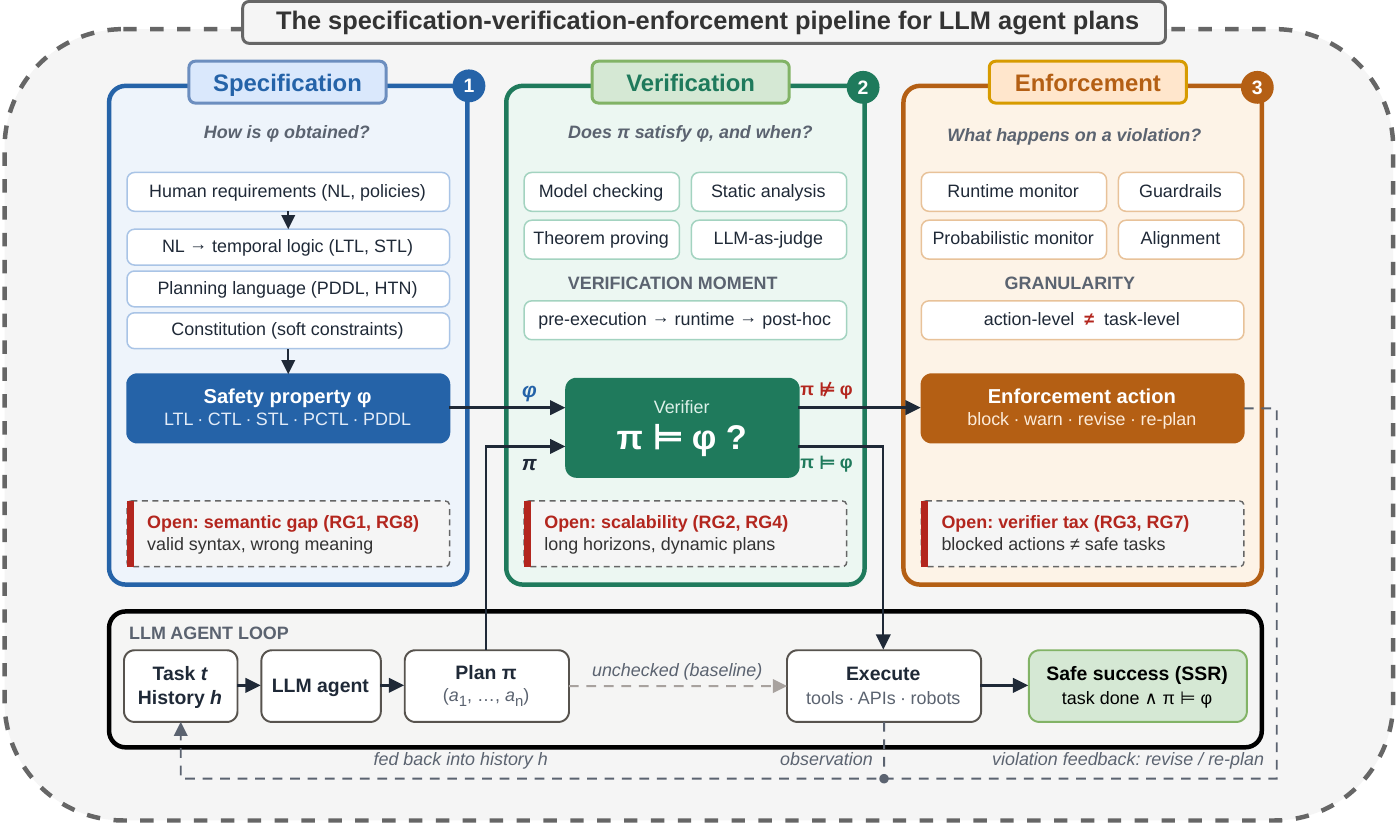}
    \caption{The specification-verification-enforcement pipeline for \gls{llm} agent plans. Given a task $t$ and history $h$, an agent $\mathcal{A}$ generates a plan $\pi$. The \emph{specification} stage (1) acquires a formal safety property $\varphi$ from human requirements; the \emph{verification} stage (2) checks whether $\pi \models \varphi$ before, during, or after execution; the \emph{enforcement} stage (3) intervenes when $\pi \not\models \varphi$. The dashed box in each stage names the research gaps (Section~\ref{sec:related:gaps}) that arise there. Only tasks completed with $\pi \models \varphi$ count toward \gls{ssr}.}
    \label{fig:pipeline}
    \Description{The specification-verification-enforcement pipeline for LLM agent plans.}
\end{figure}

We define the \emph{formal validation problem for \gls{llm} agent plans} as follows. Let $\mathcal{A}$ be an \gls{llm}-based agent that, given a task description $t$ and an observation history $h$, generates a plan $\pi = (a_1, a_2, \ldots, a_n)$ where each $a_i$ is an action (tool call, sub-task delegation, or physical action). Let $\varphi$ be a safety property expressed in a formal system $\mathcal{L}$. The verification task is to determine whether $\pi \models \varphi$, that is, whether the plan satisfies the property, before, during, or after execution, and to enforce compliance when it does not, as illustrated in Figure~\ref{fig:pipeline}. This partition reveals three tightly coupled sub-problems:
\begin{enumerate}

 \item \textbf{Specification:} How is $\varphi$ acquired from human-expressible requirements and represented in $\mathcal{L}$?

 \item \textbf{Verification:} How is $\pi \models \varphi$ determined efficiently and soundly, given that $\pi$ is generated stochastically and may be open-ended?

 \item \textbf{Enforcement:} What action is taken when $\pi \not\models \varphi$, and does that action restore task-level safety?
\end{enumerate}

Each sub-problem is independently difficult, and their composition introduces additional challenges that no single sub-problem can resolve. Figure~\ref{fig:pipeline} also marks where the main open problems of Section~\ref{sec:related:gaps} arise: the semantic gap (RG1, RG8) at specification, scalability (RG2, RG4) at verification, and the verifier tax (RG3, RG7) at enforcement.

\subsection{Scope and Inclusion Criteria}
\label{sec:intro:scope}

This survey covers:
\begin{enumerate}

 \item \gls{llm}-based agents (single or multi-agent) that produce multi-step plans, action sequences, or tool-call chains.

 \item Studies that address at least one of: plan specification, plan verification, plan enforcement, or safety monitoring.

 \item Publications from 2022 to 2026, in English, across empirical, tool, survey, and position paper categories.
\end{enumerate}

We exclude pure chatbot safety (without a planning component), neural network verification applied to non-agent settings, and works where the \gls{llm} is an incidental \gls{nlp} component in an otherwise classical system.

\subsection{Contributions}
\label{sec:intro:contrib}

This survey makes the following contributions:

\begin{enumerate}

     \item \textbf{Systematic coverage}. To the best of our knowledge, we provide the first PRISMA 2020 systematic review of the specification-verification-enforcement pipeline for \gls{llm} agent plans, identifying and synthesizing 38 studies across six databases (arXiv, ACM DL, IEEE Xplore, Semantic Scholar, Google Scholar, Preprints.org).
    
     \item \textbf{Unified taxonomy}. We introduce a three-level taxonomy (pipeline stage, verification moment, formal grounding) that organizes all included works and enables direct comparison across approaches previously described in isolation (Section~\ref{sec:taxonomy}).
    
     \item \textbf{Comparative study}. We provide a multidimensional comparison table that maps all 38 papers to pipeline stage, sub-category, verification moment (pre-execution, runtime, post-hoc), formal grounding, and venue status, and we grade the certainty of the evidence for each principal claim~\citep{Guyatt2008grade}.
    
     \item \textbf{Empirical synthesis of the verifier tax}. We aggregate evidence from enforcement and benchmark studies to characterize the relationship between action-level safety, task-level \gls{ssr}, and enforcement overhead; this relationship is the most consequential empirical finding in the corpus~\citep{Anon2026verifiertax}. This finding currently rests on a single study and requires independent replication; Section~\ref{sec:synthesis:tax} discusses its \gls{grade} certainty accordingly.
    
     \item \textbf{Structured research agenda}. We identify ten open problems (Section~\ref{sec:related:gaps}) derived from gap analysis across all five thematic clusters, with explicit pointers to the most promising technical directions for each.
     
\end{enumerate}

\subsection{Survey Organization}
\label{sec:intro:org}

We organize the remainder of this survey as follows. Section~\ref{sec:background} fixes notation and definitions; Appendix~\ref{app:background} reviews formal methods and \gls{llm} agent architectures. Section~\ref{sec:methods} details the PRISMA 2020 protocol, including eligibility criteria, search strings, the study selection flow diagram, and the data extraction procedure. Section~\ref{sec:taxonomy} presents the unified taxonomy. Section~\ref{sec:related} examines related work across five thematic clusters and highlights research gaps. Section~\ref{sec:synthesis} offers a cross-cutting synthesis, including the verifier tax analysis and \gls{grade} certainty assessment. Section~\ref{sec:agenda} summarizes research directions for each gap, with the full treatment in Appendix~\ref{app:agenda}, and analyzes verification cost under long horizons and multiple agents. Section~\ref{sec:discussion} answers the research questions stated in Section~\ref{sec:intro:problem} and discusses implications, limitations, and open challenges. Section~\ref{sec:conclusion} concludes the paper.

\section{Background and Preliminaries}
\label{sec:background}

The techniques surveyed here rest on four formal foundations: temporal logic, the language of safety properties; model checking, the decision procedure; classical planning formalisms, which supply structured task representations; and \gls{llm} agent architectures, the object of verification. Appendix~\ref{app:background} reviews all four and explains why temporal logic dominates the specification literature, why model checking is the most common algorithmic backbone, and why the stochastic nature of \glspl{llm} breaks classical assumptions. This section fixes the notation and definitions that the rest of the survey uses.

\label{sec:bg:defs}

We now formally define the core concepts that the verification pipeline operates on. These definitions serve three purposes: they fix notation for the remainder of the survey, they establish a common vocabulary across the disparate approaches we review, and they make explicit the assumptions that verification techniques rely upon. We assume that plans are finite sequences, that safety properties are expressed in temporal logic, and that \gls{ssr}, rather than action-level compliance alone, measures task-level success.

\begin{definition}[\gls{llm} Agent Plan]
  A \emph{plan} $\pi = (a_1, \ldots, a_n)$, $a_i \in \mathcal{T}$, is a finite sequence of grounded actions drawn from the agent's tool set $\mathcal{T}$, each associated with a pre-execution state $s_i$.
\end{definition}

\begin{definition}[Plan Safety Property]
  A \emph{safety property} $\varphi$ is an \gls{ltl} (or \gls{ctl}/\gls{pctl}) formula over atomic propositions derived from the action--state sequence. A plan $\pi$ \emph{satisfies} $\varphi$, written $\pi \models \varphi$, if its induced finite execution trace satisfies $\varphi$ under finite-trace semantics (LTL$_f$~\citep{DeGiacomo2013ltlf}), since a plan is a finite sequence.
\end{definition}

\begin{definition}[\acrfull{ssr}]
  Given a task set $\mathcal{T}$ and safety property $\varphi$:
  \[
    \text{SSR} = \frac{|\{t \in \mathcal{T} : \text{task completed correctly} \wedge \pi_t \models \varphi\}|}{|\mathcal{T}|}
  \]
  \gls{ssr} is the primary metric for evaluating enforcement effectiveness~\citep{Anon2026verifiertax}.
\end{definition}

A recurring challenge is the \emph{translation problem}: an \gls{llm}-based translator $\tau: \mathcal{N} \to \mathcal{F}$ from \gls{nl} to a formal language $\mathcal{F}$ (e.g., \gls{ltl}, \gls{pddl}) achieves high syntactic validity but low semantic correctness ($\tau(x) \not\equiv x$), which is the root cause of the false-assurance problem discussed in Section~\ref{sec:related:gaps}.


\section{Methodology}
\label{sec:methods}

This review follows the PRISMA 2020 guidelines~\citep{Page2021prisma} adapted for computer science literature. We conducted the review as an independent, rapid, systematic review from 20 to 29 May 2026; we did not pre-register it. We fixed the protocol (eligibility criteria, search strings, and synthesis approach) before beginning database searching.

\subsection{Eligibility Criteria}
\label{sec:methods:eligibility}

Eligibility criteria included publication years from 2022 to 2026 (excluding the pre-\gls{llm}-agent era before 2022), \gls{llm}-based agents producing multi-step plans, action sequences, or tool-call chains (excluding pure chatbot or dialogue safety absent planning), studies addressing plan specification, verification, enforcement, or safety monitoring (excluding neural network verification unless applied to agent plans), any study type (empirical, system/tool, survey, or position), and English-language full-text accessible papers (excluding conference abstracts without available content). We placed no restriction on publication type (preprint, workshop, conference, or journal); Section~\ref{sec:synthesis:grade} states how publication status is used when weighing evidence.

\subsection{Search Strategy}
\label{sec:methods:search}

Table~\ref{tab:databases} lists the six databases searched, the access method used for each, and the search window. Table~\ref{tab:searchstrings} gives the eight primary search strings verbatim, each targeting one of: formal verification of LLM agent plans, runtime verification and monitoring, temporal-logic specification of agent behavior, survey and roadmap papers on LLM agent safety, model checking with PDDL, NL-to-formal translation, the specific enforcement systems AgentSpec, AgentGuard, and VeriGuard, and formal-methods/LLM fusion roadmaps. Strings were iteratively refined during a pilot search to balance recall and precision before the protocol was fixed. Supplementary searches covered 22 named systems and authors identified during primary searching (VerifyLLM, VeriPlan, AgentVerify, ProbGuard, LogicGuard, AgentProof, NL2LTL, NL2Spec, SayCan, AgentHarm, Agent-SafetyBench, LlamaFirewall, ReAct, Toolformer, Constitutional AI, and Dafny-based LLM verification systems, among others) to confirm no additional eligible study had been missed; these 18 supplementary hits deduplicated to zero net new records against the 71 primary-search records, confirming rather than extending the primary search.

\begin{table}[ht]
\centering
\small
\caption{Databases searched.}
\label{tab:databases}
\begin{tabular}{lll}
\toprule
\textbf{Database} & \textbf{Access method} & \textbf{Date searched} \\
\midrule
arXiv (cs.AI, cs.LG, cs.SE, cs.LO) & API + WebFetch & 20--29 May 2026 \\
ACM Digital Library & \texttt{site:dl.acm.org} web search & 20--29 May 2026 \\
IEEE Xplore & \texttt{site:ieeexplore.ieee.org} web search & 20--29 May 2026 \\
Semantic Scholar & \texttt{site:semanticscholar.org} web search & 20--29 May 2026 \\
Google Scholar & Web search & 20--29 May 2026 \\
Preprints.org & \texttt{site:preprints.org} web search & 20--29 May 2026 \\
\bottomrule
\end{tabular}
\end{table}

\begin{table}[ht]
\centering
\small
\caption{Primary search strings. Year terms (e.g.\ ``2024 2025'') were searched as literal keywords against each database's default field(s) rather than applied via a dedicated publication-date filter. A keyword match on a year is a weaker proxy for publication date than a native date-range filter.}
\label{tab:searchstrings}
\begin{tabular}{cp{0.8\linewidth}}
\toprule
\textbf{\#} & \textbf{Search string} \\
\midrule
S1 & \texttt{formal verification LLM agent plans model checking 2023 2024 2025} \\
S2 & \texttt{runtime verification large language model agents safety 2023 2024} \\
S3 & \texttt{temporal logic LLM agents specification verification arXiv 2024 2025} \\
S4 & \texttt{LLM agent safety specification formal methods survey 2024 2025 arXiv} \\
S5 & \texttt{model checking AI planning PDDL formal specification agent behavior 2023 2024} \\
S6 & \texttt{natural language to formal specification LTL TLA+ translation LLM 2023 2024 2025} \\
S7 & \texttt{AgentSpec AgentGuard VeriGuard LLM agent safety enforcement 2024 2025} \\
S8 & \texttt{formal methods large language model fusion roadmap trustworthy AI agents 2024} \\
\bottomrule
\end{tabular}
\end{table}

Per-database hit counts and de-duplication counts prior to the aggregate totals in Figure~\ref{fig:prisma} were not retained; we kept only the aggregate stage totals (71 primary, 18 supplementary, 89 total, 18 duplicates), which Figure~\ref{fig:prisma} reports. We do not estimate per-string figures and treat this gap as a limitation of the review.

\subsection{Study Selection and PRISMA Flow}
\label{sec:methods:selection}

Figure~\ref{fig:prisma} gives the full PRISMA 2020 flow: 71 records from primary database search plus 18 from supplementary searches (deduplicating to zero net new records) totaling 89, 18 duplicates removed leaving 71 screened at title/abstract, 24 excluded at that stage with itemized reasons, 47 assessed at full text, 9 excluded at full text with itemized reasons, and 38 studies included. Screening proceeded in two sequential stages by a single reviewer (Section~\ref{sec:methods:irr} discusses this limitation): title/abstract screening applied the eligibility criteria of Section~\ref{sec:methods:eligibility}, retaining records with insufficient abstract information for full-text review; full-text screening then applied the same criteria under stricter methodological scrutiny. Of the 24 records excluded at title/abstract stage, 9 concerned classical AI planning without an LLM component, 8 lacked a planning or verification component (conversation safety, image-classification robustness), 4 fell outside the 2022--2026 date range, and 3 were non-English or inaccessible. Of the 9 records excluded at full-text stage, 3 were conference abstracts without accessible content, 3 concerned neural-network verification without application to agent plans, 2 used the LLM only as an incidental NLP component in an otherwise classical system, and 1 duplicated the data of an already-included study. The 38 included studies constitute the formal PRISMA corpus and form the basis for all quantitative claims, the taxonomy in Section~\ref{sec:taxonomy}, and the \gls{grade} assessments, while the larger reference list additionally includes foundational, methodological, research-agenda, and contextual comparison works that fall outside the PRISMA inclusion criteria; readers should therefore interpret all quantitative observations in this survey as about the 38 included studies only. PRISMA 2020 item 16b asks for each excluded study to be cited with its reason; because the individual full-text records were not retained (Section~\ref{sec:methods:irr}), we report the 9 full-text exclusions by category only. A completed PRISMA 2020 checklist accompanies this article as supplementary material. After the primary window closed, a targeted, non-systematic search (to 30~August~2026) identified four further 2025--2026 studies~\citep{li2026besafe,chen2025safemind,atlas2025,khan2026concurrency}. We did not apply the PRISMA protocol to them, so they are not corpus entries and appear in neither Figure~\ref{fig:prisma} nor Table~\ref{tab:taxonomy}; Sections~\ref{sec:synthesis:tax} and~\ref{sec:agenda:complexity} cite them as corroborating evidence only.

\begin{figure}[ht]
\centering
\begin{tikzpicture}[
  box/.style={rectangle, draw, rounded corners=3pt, text width=4.5cm,
              align=center, minimum height=1cm, font=\small},
  arrow/.style={->, >=stealth, thick},
  note/.style={rectangle, draw=gray, dashed, rounded corners=3pt,
               text width=4cm, align=center, minimum height=1cm, font=\footnotesize, text=gray}
]
\node[box] (id1) at (-3.5,0)
  {\textbf{Records identified\\ via database search}\\$n = 71$};
\node[box] (id2) at (3.5,0)
  {\textbf{Records from supplementary\\ named-system searches}\\$n = 18$};
\node[box] (total) at (0,-2.2)
  {\textbf{Total records identified}\\$n = 89$};
\coordinate (mergepoint) at (0,-1.0);
\draw[-, thick] (id1) -- (id1 |- mergepoint);
\draw[-, thick] (id2) -- (id2 |- mergepoint);
\draw[-, thick] (id1 |- mergepoint) -- (id2 |- mergepoint);
\draw[arrow] (mergepoint) -- (total.north);

\node[box] (dedup) at (0,-4.0)
  {\textbf{Records after duplicate removal}\\$n = 71$};
\node[note] (dupnote) at (5.5,-4.0)
  {Duplicates removed\\$n = 18$};
\draw[arrow] (total) -- (dedup);
\draw[->, dashed, gray] (dupnote.west) -- (dedup.east);

\node[box] (screen) at (0,-5.8)
  {\textbf{Records screened (title/abstract)}\\$n = 71$};
\node[box] (excl1) at (5.5,-5.8)
  {\textbf{Excluded at title/abstract}\\$n = 24$\\
  Not about \gls{llm} agents: 9\\
  No planning/verif.\ component: 8\\
  Wrong date range: 4\\
  Non-English/inaccessible: 3};
\draw[arrow] (dedup) -- (screen);
\draw[arrow] (screen) -- (excl1);

\node[box] (elig) at (0,-8.4)
  {\textbf{Full-text assessed for eligibility}\\$n = 47$};
\node[box] (excl2) at (5.5,-8.4)
  {\textbf{Excluded at full-text stage}\\$n = 9$\\
  Abstract only, no content: 3\\
  Neural network verif.\ only: 3\\
  \gls{llm} as incidental component: 2\\
  Duplicate data: 1};
\draw[arrow] (screen) -- (elig);
\draw[arrow] (elig) -- (excl2);

\node[box, fill=gray!10] (incl) at (0,-10.4)
  {\textbf{Studies included in synthesis}\\$n = 38$};
\draw[arrow] (elig) -- (incl);
\end{tikzpicture}
\caption{PRISMA 2020 flow diagram for study selection.}
\Description{PRISMA 2020 flow diagram. Database searches yielded 71 records and supplementary named-system searches 18, for 89 in total; removing 18 duplicates left 71 records screened by title and abstract. Of these, 24 were excluded (not about LLM agents, 9; no planning or verification component, 8; wrong date range, 4; non-English or inaccessible, 3), and 47 were assessed in full text. Nine were excluded at full text (abstract only, 3; neural network verification only, 3; LLM as an incidental component, 2; duplicate data, 1), leaving 38 studies included in the synthesis.}
\label{fig:prisma}
\end{figure}

\subsection{Data Extraction}
\label{sec:methods:extraction}

For each included study we extracted, using a standardized template, five field groups: (i) \emph{bibliographic} fields (authors, year, title, venue, arXiv ID or DOI); (ii) \emph{classification} fields (pipeline stage SPEC/VERIF/ENF, sub-category, formal method(s) used); (iii) \emph{technical} fields (key contribution, verification moment Pre/RT/PH, formal grounding); (iv) \emph{evaluation} fields (evaluation method, benchmark(s), key quantitative results, disclosed limitations); and (v) \emph{quality} fields, assessed qualitatively per study against four bias domains: construct validity (are the tested safety properties aligned with real-world agent risks?), internal validity (are experimental controls and baselines adequate?), generalizability (are evaluations limited to toy domains?), and reporting completeness (are failure modes disclosed?). We use these domains because the Cochrane RoB 2 tool does not directly apply to a corpus dominated by system/tool papers and surveys rather than randomized controlled experiments. We did not extract the funding sources of the included studies, since the synthesis concerns the technical content of each study rather than its provenance. Data were extracted directly from each source paper's text, abstract, or accessible supplementary material; where quantitative results were reported in incompatible units or metrics, they are recorded descriptively rather than numerically combined. The decision rule for studies spanning more than one Level-1 pipeline stage (Section~\ref{sec:taxonomy:l1}) is stated in Section~\ref{sec:taxonomy}: a study is labeled under more than one stage only when it both proposes a specification artifact and separately implements a mechanism that enforces or verifies it, the case for Constitutional~AI (SPEC-AL and ENF-ALIGN). Borderline eligibility decisions were resolved conservatively (retain if in doubt); Section~\ref{sec:discussion:threats} discusses the effect of this rule on inclusion counts. The completed extraction spreadsheet (one row per included study, the fields above as columns, built directly from Table~\ref{tab:taxonomy} and the bibliography) is deposited alongside the other materials described in Section~\ref{sec:methods:availability}.

Applying the four quality domains above to each of the 38 included studies individually yields four strata: 6 studies at \emph{low} risk (large, multi-model, multi-domain benchmark papers with clear methodology and disclosed limitations, e.g.\ Agent-SafetyBench~\citep{Zhang2024agentsafety}, AgentHarm~\citep{Andriushchenko2024agentharm}, AgentBench~\citep{Liu2023agentbench}, and PlanBench~\citep{Valmeekam2023planbench}); 14 studies at \emph{moderate} risk (peer-reviewed single-study system/tool papers with clear methodology but curated benchmarks, limited domain coverage, and no independent replication, e.g.\ NL2LTL~\citep{He2023nl2ltl}, AgentSpec~\citep{Wang2025agentspec}, and the verifier-tax finding~\citep{Anon2026verifiertax}); 12 studies at \emph{high} risk (preprints or short/early-stage tracks with thin single-domain evaluation, unquantified effectiveness claims, or LLM-as-Judge evaluation without ground truth, e.g.\ the LLM-based embodied-task plan verification study~\citep{Li2025planverify}, AgentVerify~\citep{Anon2026agentverify}, and Constitutional~\gls{ai}~\citep{Bai2022constitutional}); and 6 studies classified \emph{not applicable} (foundational architecture, narrative survey, or position papers with no falsifiable empirical verification claim to rate, e.g.\ Toolformer~\citep{Schick2023toolformer}, Chain-of-Thought~\citep{Wei2022cot}, and ReAct~\citep{Yao2023react}). The per-study assignment is available in the data deposit (Section~\ref{sec:methods:availability}); for four studies (Agent-SafetyBench, AgentHarm, AgentBench, and the \gls{llm}-based plan verification study) the sheet records the overall tier only, not a tier per domain. Three risks apply across the corpus as a whole rather than to individual studies: benchmark contamination (LLMs may have encountered evaluation tasks during pre-training), an evaluation-deployment gap (most studies evaluate on static or sandboxed tasks, whereas real agentic deployments are open-ended), and publication bias toward positive results, discussed further in Section~\ref{sec:synthesis:grade}.

\subsection{Dual Screening and Inter-Rater Reliability}
\label{sec:methods:irr}

A single reviewer (P.D.) performed the primary screening; Section~\ref{sec:discussion:threats} states this as a threat to internal validity. To validate it, we repeated the eight search strings (Table~\ref{tab:searchstrings}) on the six databases on 22--23~September~2026. Because the 71 title/abstract and 47 full-text records of the primary screening were not retained individually, the check uses the 37 records that this repeated search returned. Three authors screened all 37 at title/abstract level against the eligibility criteria of Section~\ref{sec:methods:eligibility}: the primary reviewer and two co-authors (L.C. and E.N.) who had no part in the primary screening. The two co-authors screened blind to each other's decisions, to the primary reviewer's decisions, and to the 38-study included list; the primary reviewer screened the repeated set without consulting that list.

Thirty-two of the 37 records received unanimous decisions, and Fleiss'~$\kappa$ across the three raters is 0.81 (bootstrap 95\% CI [0.63, 0.96]). Pairwise, raw agreement was \SI{89.2}{\percent} between P.D. and L.C. (33/37; Cohen's~$\kappa = 0.77$, 95\% CI [0.56, 0.98]), \SI{86.5}{\percent} between P.D. and E.N. (32/37; $\kappa = 0.71$, [0.47, 0.95]), and \SI{97.3}{\percent} between L.C. and E.N. (36/37; $\kappa = 0.94$, [0.83, 1.00]). Inclusion prevalence was near balance (22 to 24 includes per rater), so these $\kappa$ values do not suffer from the prevalence paradox that affects skewed samples. We resolved the five disagreements by majority vote. In four of them the two co-authors agreed with each other against the primary reviewer, and three of those four were records the primary reviewer included and both co-authors excluded: CASTL, a tool-orchestration safety-evaluation paper also named AgentGuard, and an MCP guardrail architecture paper. Both co-authors judged the agent-plan component of these papers peripheral. We read this pattern as the retain-if-in-doubt rule (Section~\ref{sec:methods:selection}) at work, and it is consistent with the inflation risk that Section~\ref{sec:discussion:threats} names. The fourth, a general survey of LLM safeguards, went the other way. The fifth, RV4Chatbot, split the two co-authors.

Thirteen of the 37 records are in the 38-study corpus. All three raters included ten of them (AgentSpec, VeriGuard, AgentProof, the LLM+FM roadmap, LogicGuard, VerifyLLM, the LLM+FM plan-verification case study, Agent-C, VeriPlan, and ProbGuard, which the repeated search returned under its earlier arXiv title, Pro2Guard). All three, the primary reviewer included, excluded the other three: NL2LTL~\citep{He2023nl2ltl}, NL2Spec~\citep{Cosler2023nl2spec}, and NL-to-LTL/HSD~\citep{Anon2025nl2ltl_hsd}. These papers translate natural language into temporal logic in general, with no LLM agent producing plans, and the raters judged them outside the second eligibility criterion as written. We keep them in the corpus, since the SPEC-NL sub-category (Section~\ref{sec:taxonomy}) rests on them, but their inclusion is not unambiguous: claims that draw on them, including the syntactic-correctness row of Table~\ref{tab:grade}, describe general NL-to-temporal-logic translation rather than agent-plan specification specifically. In the other direction, all three raters included 11 records that are not in the corpus. Ten of them predate the primary search window (20--29~May~2026) and are gaps in the primary search: GuardAgent (arXiv:2406.09187), AGrail (arXiv:2502.11448), ShieldAgent (arXiv:2503.22738), PAT-Agent (ASE~2025), SafePilot, AgentGuard (arXiv:2509.23864, September~2025), Watchdogs and Oracles (arXiv:2511.14435, November~2025), SpecMAS (NeurIPS~2025), an end-to-end agentic PDDL planning framework (arXiv:2512.09629, December~2025), and an AgenticOps runtime-verification framework (IJRSI, December~2025). The eleventh, an NL-to-strategic-temporal-specification translator (arXiv:2606.30441), first appeared on 29~June~2026, after the window closed. We report all eleven as a stated limitation rather than adding them to the formal corpus outside the protocol, the same principle applied to the August~2026 targeted search (Section~\ref{sec:methods:selection}).

This check has four limits. It covers title/abstract screening only; no second rater repeated full-text screening, data extraction, or the risk-of-bias tiers of Section~\ref{sec:methods:extraction}. The two additional raters are co-authors, not external reviewers. The repeated search returned 37 records against the primary search's 71 and surfaced only 13 of the 38 corpus studies, so it measures screening consistency on a comparable record set, not the reproducibility of the primary search. Finally, the primary reviewer's exclusion of the three NL-to-LTL papers in this check reverses that reviewer's inclusion decision on them, which is a within-rater inconsistency the check exposed rather than resolved. The three-rater screening sheet is part of the data deposit (Section~\ref{sec:methods:availability}).

\subsection{Data and Materials Availability}
\label{sec:methods:availability}

The classification data underlying Table~\ref{tab:taxonomy}, the \gls{grade} worksheets underlying Table~\ref{tab:grade}, the per-study risk-of-bias rating sheet underlying Section~\ref{sec:methods:extraction}, the title/abstract and full-text screening log (at the aggregate level; Sections~\ref{sec:methods:search} and~\ref{sec:methods:irr} explain why), and the three-rater screening sheet for the reliability check of Section~\ref{sec:methods:irr} are deposited at Zenodo, DOI: \url{https://doi.org/10.5281/zenodo.23376335}. We do not deposit the per-database hit counts underlying Section~\ref{sec:methods:search}'s search-strategy tables, since those counts were not retained.

\section{Taxonomy and Classification System}
\label{sec:taxonomy}

We introduce a three-level taxonomy (Sections~\ref{sec:taxonomy:l1} to~\ref{sec:taxonomy:l3}) classifying all 38 included studies along three orthogonal questions: \emph{which pipeline stage} a study belongs to (Level~1), \emph{when} verification occurs relative to execution (Level~2, verification moment), and \emph{how} verification is performed (Level~3, formal grounding). The three dimensions correlate in practice rather than varying fully independently: SPEC studies have no verification moment of their own (Table~\ref{tab:taxonomy} marks the five NL-to-formal rows Pre, since their formulas are consumed before execution, and two rows N/A), with SayCan a labeled exception (its affordance-function scoring suppresses low-affordance candidate actions at runtime, so it carries a genuine RT moment despite its SPEC-PL stage); ENF studies are predominantly runtime by definition, with VerifyLLM a labeled exception (a pre-execution intervention, still classified ENF because it blocks rather than merely reports); and we distinguish VERIF (observes and reports a violation) from ENF-RT (observes and intervenes), even though both are commonly called ``runtime monitoring'' in their source papers. Multi-label classification follows one rule, applied consistently: a study is labeled under more than one Level-1 stage only when it both proposes a specification artifact and separately implements a mechanism that enforces or verifies it; Constitutional~AI is the one such case in this corpus (SPEC-AL and ENF-ALIGN), since it proposes a constitution and separately trains a policy to conform to it. Appendix~\ref{app:taxonomies} compares the taxonomy with three earlier classifications.

\subsection{Level 1 -- Pipeline Stage}
\label{sec:taxonomy:l1}

\subsubsection{Specification (SPEC)}

Specification techniques handle how safety properties are acquired and expressed. They are the entry point of the pipeline; verification techniques consume their output. We identify three sub-categories:

\begin{itemize}

 \item \textbf{SPEC-\gls{nl}}: \gls{nl} to formal specification translation. The input is a human-expressed requirement; the output is a formula in a formal notation. Examples: NL2LTL~\citep{He2023nl2ltl}, NL2Spec~\citep{Cosler2023nl2spec}, \gls{nl}-to-\gls{ltl}/\gls{hsd}~\citep{Anon2025nl2ltl_hsd}, \gls{nl}-to-\gls{stl}~\citep{Fang2025nl2stl,Anon2026clarifystl}.

 \item \textbf{SPEC-PL}: Planning language specification. The input is a task description; the output is a \gls{pddl} domain/problem or \gls{htn} description. Examples: \glspl{llm} as Planning Formalizers~\citep{Anon2025pddl_survey}, SayCan~\citep{Ahn2022saycan}.

 \item \textbf{SPEC-AL}: Alignment-based specification. This approach encodes safety requirements as soft constraints, using training objectives or constitutions rather than formal formulas. Example: Constitutional \gls{ai}~\citep{Bai2022constitutional}, whose 58-principle constitution is a specification artifact separate from the \gls{rlaif}/\gls{slaif} training that enforces it. Agent Safety Alignment via \gls{rl}~\citep{Anon2025rlsafety} shares the alignment-training mechanism but does not articulate a comparable standalone specification artifact, so it is classified ENF-ALIGN only (Section~\ref{sec:taxonomy:l1}).
\end{itemize}

\subsubsection{Verification (VERIF)}

Verification techniques determine whether a plan or plan fragment satisfies a specification. We identify four sub-categories:

\begin{itemize}

 \item \textbf{VERIF-MC}: Model checking. The approach translates plans or plan representations into Kripke structures or FSMs, on which model-checking algorithms check \gls{ltl}/\gls{ctl} properties. Examples: Ramani et al.~\citep{Ramani2025bridging}, AgentVerify~\citep{Anon2026agentverify}, VeriPlan~\citep{Anon2025veriplan}, Logic-based MAS Verification~\citep{Anon2026logicmas}.

 \item \textbf{VERIF-SA}: Static analysis. Pre-execution structural checks on agent workflow graphs or code. Examples: AgentProof~\citep{Li2026agentproof}, Dafny-as-IL~\citep{Anon2025dafny}, DafnyPro~\citep{Anon2026dafnypro}.

 \item \textbf{VERIF-TP}: Theorem proving. Deductive verification against formal contracts. Examples: AutoRocq~\citep{NUS2025autorocq}, VeriGuard~\citep{Miculicich2025veriguard}.

 \item \textbf{VERIF-\gls{llm}}: \gls{llm}-assisted verification. An \gls{llm} serves as a critic or judge, iteratively checking plan quality without a formal prover. Examples: Self-Refine~\citep{Madaan2023selfrefine}, Plan Verification via \gls{llm}-as-Judge~\citep{Li2025planverify}.
\end{itemize}

\subsubsection{Enforcement (ENF)}

Enforcement techniques define what happens when verification finds a violation or anticipates one. We identify four sub-categories:

\begin{itemize}

 \item \textbf{ENF-RT}: Runtime monitoring and blocking: online monitors intercept unsafe actions before execution. Examples: AgentSpec~\citep{Wang2025agentspec}, Agent-C~\citep{Kamath2025agentc}, VerifyLLM~\citep{Grigorev2025verifyllm}, LogicGuard~\citep{Gokhale2025logicguard}.

 \item \textbf{ENF-PROB}: Probabilistic monitoring. Stochastic models forecast subsequent violations and trigger early warnings. Examples: ProbGuard~\citep{Wang2025probguard}.

 \item \textbf{ENF-SYS}: System-level guardrails. Multi-layer security infrastructure operating on prompts, reasoning traces, and generated code. Example: LlamaFirewall~\citep{MetaAI2025llamafirewall}.

 \item \textbf{ENF-ALIGN}: Alignment-based enforcement. Model training enforces safety rather than runtime checks. Examples: Constitutional \gls{ai}~\citep{Bai2022constitutional}, Agent Safety Alignment via \gls{rl}~\citep{Anon2025rlsafety}.
\end{itemize}

\subsection{Level 2 -- Verification Moment}
\label{sec:taxonomy:l2}

The literature distinguishes three verification moments, each reflecting a different trade-off between the strength of the safety guarantee and the practical constraints of plan availability and executability.

\begin{itemize}
    \item \textbf{Pre-execution verification}: The system checks the plan in full before executing any action. This strategy permits rejection or revision before irreversible effects occur, but requires that the complete plan be available. Pre-execution verification appears in AgentProof~\citep{Li2026agentproof}, the offline phase of VeriGuard~\citep{Miculicich2025veriguard}, and VerifyLLM~\citep{Grigorev2025verifyllm}. 
    
    \item \textbf{Runtime verification}: The system checks each action before executing it (or at most $k$ steps in advance). This approach handles dynamically generated plans but cannot verify future actions that the system has not yet generated. Runtime verification appears in AgentSpec~\citep{Wang2025agentspec}, Agent-C~\citep{Kamath2025agentc}, LogicGuard~\citep{Gokhale2025logicguard}, and ProbGuard~\citep{Wang2025probguard}. 
    
    \item \textbf{Post-hoc verification}: The system analyzes the complete execution trace after the task concludes. This approach is useful for auditing and offline learning, but cannot prevent harm during execution. Post-hoc verification appears in Self-Refine~\citep{Madaan2023selfrefine} (in iterative mode) and \gls{llm}-as-Judge~\citep{Li2025planverify}.

\end{itemize}

\subsection{Level 3 -- Formal Grounding}
\label{sec:taxonomy:l3}

The included studies rely on formal and heuristic backbones. Temporal Logic uses \gls{ltl}, \gls{ctl}, \gls{stl}, or \gls{pctl} as the specification and verification language, supporting sequential, timing, and probabilistic safety properties. It is the most common grounding in the corpus (Section~\ref{sec:taxonomy:obs}). 

\begin{itemize}
    \item \textbf{Classical planning (CP)}: Employs \gls{pddl} or \gls{htn} as the plan representation and verification target, enabling soundness guarantees through established plan validation tools. It appears in \gls{pddl}-survey work~\citep{Anon2025pddl_survey} and SayCan~\citep{Ahn2022saycan}. 
    
    \item \textbf{Theorem proving (TP)}: Relies on Coq/Rocq or Dafny as the verification backend, providing complete correctness proofs for programs and policies. It offers the highest assurance level but is currently limited to code-based agents, as in AutoRocq~\citep{NUS2025autorocq}, Dafny-as-IL~\citep{Anon2025dafny}, DafnyPro~\citep{Anon2026dafnypro}, and VeriGuard~\citep{Miculicich2025veriguard}.

    \item \textbf{Graph and automata methods (GA)}: Use \gls{dfa}, \gls{fsm}, or directed graph product constructions for efficient workflow-level static verification. They appear in AgentProof~\citep{Li2026agentproof} and AgentVerify~\citep{Anon2026agentverify}. 
    
    \item \textbf{Probabilistic methods (PR)}: Draw on \gls{dtmc}, \gls{mdp}, or \gls{pac}-learning bounds, suiting stochastic agents where deterministic assurances remain unavailable. ProbGuard~\citep{Wang2025probguard} illustrates this approach. 
    
    \item \textbf{Heuristic and mixed approaches (HH)}: Encompass methods without a single formal mathematical backing, including: (i) \gls{llm}-as-judge criticism (Self-Refine~\citep{Madaan2023selfrefine}, Plan Verification via \gls{llm}~\citep{Li2025planverify}); (ii) \gls{rl} reward shaping and \gls{rlaif}/\gls{slaif} training (Constitutional \gls{ai}~\citep{Bai2022constitutional}, Agent Safety Alignment via \gls{rl}~\citep{Anon2025rlsafety}); and (iii) multi-heuristic system pipelines (LlamaFirewall~\citep{MetaAI2025llamafirewall}). These subtypes differ in scalability, soundness properties, and deployment requirements; readers should not treat HH as a homogeneous category.

\end{itemize}

\subsection{Full Classification Table}
\label{sec:taxonomy:table}

Table~\ref{tab:taxonomy} classifies all 38 included studies along the three taxonomy dimensions, grouped by pipeline stage. For each study, the table records the sub-category, verification moment (Pre = pre-execution, RT = runtime, PH = post-hoc), formal grounding, and venue status. The table does not grade individual studies: \gls{grade} rates the evidence for a claim, so certainty ratings appear per claim in Table~\ref{tab:grade}. Column abbreviations. Sub-cat: Arch. = foundational architecture, Survey = narrative survey/position paper, NL = NL-to-formal, PL = planning, AL = alignment, MC = model checking, SA = static analysis, TP = theorem proving, LV = \gls{llm}-verified, Bench = benchmark-based outcome evaluation, RT-mon = runtime monitoring, PR-mon = probabilistic monitoring, SYS = system guardrail. Venue: PR = peer-reviewed proceedings or journal (workshop venues so noted); PP = preprint only, independently checked against Crossref, ACL Anthology, IEEE Xplore, ACM DL, and official conference/workshop programs at the time of writing (see Section~\ref{sec:synthesis:grade}). Three rows carry a venue-status caveat: $^{\ddagger}$the study has an active OpenReview/workshop submission whose archival status could not be confirmed from public sources, and is provisionally marked PP pending confirmation. The six rows marked $^{*}$ establish a foundational representation format (e.g.\ inline tool-call annotation, chain-of-thought reasoning traces) or survey the field narratively, without proposing or implementing a SPEC, VERIF, or ENF technique themselves; later work in the corpus builds on or verifies the representations they establish (Section~\ref{sec:related}). Bench rows (Agent-SafetyBench, AgentHarm, AgentBench, PlanBench) denote benchmark-based outcome evaluation against a safety or task-success metric, distinct from a formal proof obligation; see Section~\ref{sec:related:bench} for the scope of this distinction.

\begingroup
\small
\begin{longtable}{lllllll}
\caption{Classification of all 38 included studies under the proposed taxonomy, grouped by pipeline stage.}
\label{tab:taxonomy}\\
\toprule
\textbf{System / Paper} & \textbf{Year} & \textbf{Venue} &
\textbf{Stage} & \textbf{Sub-cat} & \textbf{Moment} &
\textbf{Grounding} \\
\midrule
\endfirsthead
\multicolumn{7}{l}{\small\textit{Table~\ref{tab:taxonomy} continued from previous page}} \\
\toprule
\textbf{System / Paper} & \textbf{Year} & \textbf{Venue} &
\textbf{Stage} & \textbf{Sub-cat} & \textbf{Moment} &
\textbf{Grounding} \\
\midrule
\endhead
\midrule
\multicolumn{7}{r}{\small\textit{continued on next page}} \\
\endfoot
\bottomrule
\endlastfoot
\multicolumn{7}{l}{\textit{Foundational and Survey Works (FS) -- outside the three-stage taxonomy}$^{*}$} \\
\midrule
Toolformer~\citep{Schick2023toolformer}         & 2023 & PR (NeurIPS) & FS & Arch.  & N/A & -- \\
Chain-of-Thought~\citep{Wei2022cot}             & 2022 & PR (NeurIPS) & FS & Arch.  & N/A & -- \\
Planning Survey~\citep{Huang2024planning}       & 2024 & PP & FS & Survey & N/A & -- \\
LLM+FM Roadmap~\citep{Zhang2024roadmap}         & 2024 & PP & FS & Survey & N/A & -- \\
ReAct~\citep{Yao2023react}                       & 2023 & PR (ICLR) & FS & Arch.  & N/A & -- \\
Full Stack Safety~\citep{Anon2025fullstack}      & 2025 & PP & FS & Survey & N/A & -- \\
\midrule
\multicolumn{7}{l}{\textit{Specification (SPEC)}} \\
\midrule
NL2LTL~\citep{He2023nl2ltl}                    & 2023 & PR (AAAI demo) & SPEC & NL     & Pre & TL \\
NL2Spec~\citep{Cosler2023nl2spec}               & 2023 & PR (CAV) & SPEC & NL     & Pre & TL \\
NL-to-LTL/HSD~\citep{Anon2025nl2ltl_hsd}       & 2025 & PR (ASE) & SPEC & NL     & Pre & TL \\
NL-to-STL~\citep{Fang2025nl2stl}               & 2025 & PR (ACL Find.) & SPEC & NL     & Pre & TL \\
ClarifySTL~\citep{Anon2026clarifystl}           & 2026 & PP & SPEC & NL     & Pre & TL \\
\gls{pddl} Survey~\citep{Anon2025pddl_survey}  & 2025 & PR (ACL Find.) & SPEC & PL     & N/A & CP \\
SayCan~\citep{Ahn2022saycan}                    & 2022 & PR (CoRL) & SPEC & PL     & RT  & CP \\
Constitutional \gls{ai}~\citep{Bai2022constitutional} & 2022 & PP & SPEC & AL & N/A & HH \\
\midrule
\multicolumn{7}{l}{\textit{Verification (VERIF)}} \\
\midrule
Bridging \gls{llm}+FM~\citep{Ramani2025bridging}  & 2025 & PR (AgenticSE@ASE, workshop) & VERIF & MC    & Pre & TL \\
AgentVerify~\citep{Anon2026agentverify}           & 2026 & PP & VERIF & MC    & Pre & TL+GA \\
Logic-based MAS~\citep{Anon2026logicmas}          & 2026 & PP & VERIF & MC    & Pre & TL \\
VeriPlan~\citep{Anon2025veriplan}                 & 2025 & PR (CHI) & VERIF & MC    & Pre & TL \\
AgentProof~\citep{Li2026agentproof}               & 2026 & PP & VERIF & SA    & Pre & GA \\
Dafny-as-IL~\citep{Anon2025dafny}                & 2025 & PR (Dafny@POPL, workshop) & VERIF & SA    & Pre & TP-prov \\
DafnyPro~\citep{Anon2026dafnypro}                & 2026 & PR (Dafny@POPL, workshop) & VERIF & SA    & Pre & TP-prov \\
VeriGuard~\citep{Miculicich2025veriguard}         & 2025 & PP & VERIF & TP    & Pre & TP-prov \\
AutoRocq~\citep{NUS2025autorocq}                 & 2025 & PR (PACMSE/FSE) & VERIF & TP    & Pre & TP-prov \\
Self-Refine~\citep{Madaan2023selfrefine}          & 2023 & PR (NeurIPS) & VERIF & LV    & PH  & HH \\
Plan Verif.\ \gls{llm}~\citep{Li2025planverify}  & 2025 & PP$^{\ddagger}$ & VERIF & LV    & PH  & HH \\
Agent-SafetyBench~\citep{Zhang2024agentsafety}    & 2024 & PP & VERIF & Bench & PH  & HH \\
AgentHarm~\citep{Andriushchenko2024agentharm}     & 2024 & PR (ICLR) & VERIF & Bench & PH  & HH \\
AgentBench~\citep{Liu2023agentbench}              & 2023 & PR (ICLR) & VERIF & Bench & PH  & HH \\
PlanBench~\citep{Valmeekam2023planbench}          & 2023 & PR (NeurIPS D\&B) & VERIF & Bench & PH  & CP \\
\midrule
\multicolumn{7}{l}{\textit{Enforcement (ENF)}} \\
\midrule
Agent-C~\citep{Kamath2025agentc}                & 2025 & PP$^{\ddagger}$ & ENF & RT-mon & RT  & TL \\
AgentSpec~\citep{Wang2025agentspec}              & 2025 & PR (ICSE) & ENF & RT-mon & RT  & GA \\
LogicGuard~\citep{Gokhale2025logicguard}         & 2025 & PP$^{\ddagger}$ & ENF & RT-mon & RT  & TL \\
VerifyLLM~\citep{Grigorev2025verifyllm}          & 2025 & PR (IROS) & ENF & RT-mon & Pre & TL \\
Verifier Tax~\citep{Anon2026verifiertax}         & 2026 & PR (CAIS) & ENF & RT-mon & RT  & HH \\
Safe Tool Use~\citep{Anon2025toolusetool}        & 2025 & PR (ICSE NIER) & ENF & RT-mon & RT  & TL \\
ProbGuard~\citep{Wang2025probguard}              & 2025 & PR (ASE, accepted) & ENF & PR-mon & RT  & PR \\
LlamaFirewall~\citep{MetaAI2025llamafirewall}    & 2025 & PP & ENF & SYS    & RT  & HH+GA \\
Constitutional \gls{ai}~\citep{Bai2022constitutional} & 2022 & PP & ENF & AL  & N/A & HH \\
\gls{rl} Safety~\citep{Anon2025rlsafety}        & 2025 & PP & ENF & AL     & N/A & HH \\
\end{longtable}
\endgroup

\subsection{Taxonomy-Derived Observations}
\label{sec:taxonomy:obs}

Inspecting Table~\ref{tab:taxonomy} reveals five structural patterns that recur throughout the synthesis in Section~\ref{sec:synthesis} and the discussion in Section~\ref{sec:discussion}.

\begin{enumerate}

    \item \textbf{Temporal Logic dominance}: Temporal Logic (\gls{ltl}/\gls{ctl}/\gls{stl}) is the formal grounding for 13 of 38 studies (\SI{34}{\percent}), showing its suitability for expressing sequential agent safety properties. Graph/automata and theorem-proving approaches each account for approximately \SI{11}{\percent} of studies (4 papers each). Probabilistic grounding (\gls{dtmc}/\gls{mdp}) is underrepresented (one study), a gap given the stochastic nature of \gls{llm} agents.

    \item \textbf{Pre-execution verification is the most common moment; runtime verification concentrates in enforcement}: Of 38 studies, 15 (\SI{39}{\percent}) operate pre-execution, but the majority of these require finite, pre-specified workflow graphs. Runtime verification accounts for 8 studies (\SI{21}{\percent}), seven of them enforcement systems. Post-hoc verification accounts for 6 studies (\SI{16}{\percent}) and is predominantly \gls{llm}-assisted. The remaining 9 studies (\SI{24}{\percent}) have no defined verification moment (N/A in Table~\ref{tab:taxonomy}): the 6 foundational-architecture and survey works outside the three-stage taxonomy, plus PDDL Survey, Constitutional AI, and RL Safety, whose SPEC/ENF contributions are moment-agnostic by construction.
    
    \item \textbf{Specification and enforcement are relatively underserved}: Only 8 studies focus primarily on specification (\SI{21}{\percent}), and 10 on enforcement (\SI{26}{\percent}); one paper (Constitutional AI~\citep{Bai2022constitutional}) is classified under both SPEC and ENF. The verification cluster is the largest (15 studies, \SI{39}{\percent}). The remaining 6 studies (\SI{16}{\percent}) are foundational architecture or narrative survey works that establish representations or characterize the field without proposing or implementing a SPEC, VERIF, or ENF technique themselves, and so sit outside the three-stage taxonomy (Table~\ref{tab:taxonomy}, Section~\ref{sec:taxonomy:l1}). This unevenness reflects the field's origins in formal methods, where verifying a given specification is the core problem; specification acquisition and post-violation enforcement are comparatively understudied.
    
    \item \textbf{Heuristic/hybrid grounding accounts for about a quarter of included work}: 9 studies (\SI{24}{\percent}) rely on heuristic or hybrid grounding, primarily \gls{llm}-as-judge, \gls{rl}, or soft alignment. These approaches scale well but provide no formal soundness guarantees. The tension between formal soundness (low scalability) and empirical scalability (no soundness) is the central unresolved tradeoff in the field.
    
    \item \textbf{\gls{grade} certainty is predominantly Very Low}: Of the nine principal claims in Table~\ref{tab:grade}, two reach Moderate, seven are Very Low, and none reach High. Narrow domain scope, single-study findings for most key claims, the absence of randomized evaluations, and high heterogeneity in benchmarks limit the field's certainty.

\end{enumerate}


\section{Related Work}
\label{sec:related}

The intersection of \glspl{llm} agents and formal validation is an emerging, rapidly evolving field. We organize the related work into five thematic clusters that mirror the pipeline from plan representation to safety enforcement: (i)~foundational agent architectures and plan representations, (ii)~specification techniques, (iii)~verification approaches, (iv)~enforcement and correction mechanisms, and (v)~benchmarks and evaluation. Within each cluster, we compare the principal contributions, identify their limitations, and highlight the gaps that motivate the present survey.

\subsection{Comparison with Prior Surveys}
\label{sec:related:prior}

Table~\ref{tab:related_surveys} compares this survey with three prior works that most closely overlap its scope. No prior survey applies a systematic search protocol to the complete specification-verification-enforcement pipeline, provides \gls{grade} certainty grading, or synthesizes the verifier tax as a field-level empirical finding.

The four 2025--2026 studies that a targeted search found after the primary window (Section~\ref{sec:methods:selection}) are corroborating evidence only. None of them, and none of the three prior surveys in Table~\ref{tab:related_surveys}, applies a PRISMA 2020 protocol, \gls{grade} certainty grading, and a field-level synthesis of the verifier tax to the specification-verification-enforcement pipeline as a whole.

\begin{table}[t]
\centering
\small
\caption{Comparison with related surveys. Type: SLR = Systematic Literature Review; NS = Narrative Survey. Pipeline: S = Specification, V = Verification, E = Enforcement.}
\label{tab:related_surveys}
\begin{tabular}{ccccccc}
\toprule
\textbf{Survey} & \textbf{Type} & \textbf{Scope} & \textbf{Method} & \textbf{FM depth} & \textbf{Pipeline} & \textbf{Evidence} \\
\midrule
Zhang et al.~\citep{Zhang2024roadmap} & NS & FM--\gls{llm} integration & Narrative & Mod. & S, V & None \\
Full-Stack Safety~\citep{Anon2025fullstack} & NS & Full \gls{llm} safety & Narrative & Low & Partial & None \\
Huang et al.~\citep{Huang2024planning} & NS & \gls{llm} planning & Narrative & None & S only & None \\
\midrule\textbf{This survey} & \textbf{SLR} & \textbf{Spec-Verif-Enf} & \textbf{PRISMA 2020} & \textbf{Deep} & \textbf{S,V,E} & \textbf{\gls{grade}} \\
\bottomrule
\end{tabular}
\end{table}

\subsection{Foundational Agent Architectures and Plan Representations}
\label{sec:related:arch}

The representation of an agent's plan determines much of the tractability of formal verification. Analysis of the 38 included studies reveals three representation paradigms that have emerged from the \gls{llm}-agent literature, each with distinct properties from a verification standpoint.

First, \textbf{trace-based representations} (ReAct, \gls{cot}) use an \gls{llm} to alternate between \gls{nl} ``thought'' steps and external ``action'' calls, resulting in a sequence of (thought, action, observation) triples~\citep{Yao2023react}. This approach makes reasoning explicit and each plan step easy to follow, producing an execution trace suitable for \gls{ltl} model checking~\citep{Ramani2025bridging,Grigorev2025verifyllm}. In larger \glspl{llm}, \gls{cot} prompting~\citep{Wei2022cot} generates multi-step reasoning traces whose precondition--action--effect structure resembles STRIPS operator schemas~\citep{Anon2025pddl_survey}, allowing partial translation to classical planning formats. Both representations share one limitation: plan correctness is only checked \emph{after} the model has generated the full thought-action trace.

Second, \textbf{tool-augmented representations} (Toolformer, SayCan) train \glspl{llm} to self-annotate API calls inline with \gls{nl} text~\citep{Schick2023toolformer}. The resulting hybrid representation, \gls{nl} interleaved with embedded API call slots, is hard to extract formally because the boundary between specification and execution is implicit. \gls{llm} planning has been grounded in robotic affordance functions, scoring candidate actions using both language model probabilities and a skill-value function~\citep{Ahn2022saycan}. SayCan provides an implicit safety filter (the system suppresses low-affordance actions) but lacks formal guarantees; we best understand it as a learned precursor to explicit enforcement.

Third, \textbf{graph-based representations} (LangGraph, CrewAI, AutoGen) represent plans as directed state graphs with typed nodes (agents, tools, and control structures) and typed edges (data movement and control flow). This representation is directly amenable to static graph analysis and product-automaton model checking~\citep{Li2026agentproof}. A prior taxonomy classifies \gls{llm} planning into task decomposition, plan selection, external module use, and reflection/memory, each corresponding to a different verification entry point~\citep{Huang2024planning}.

Trace-based representations are the most widely studied for runtime verification, whereas graph-based representations are the most amenable to pre-deployment static analysis. Neither paradigm simultaneously supports complete pre-execution verification and arbitrary open-ended planning. The core representational gap is the absence of a \emph{canonical intermediate representation} that faithfully retains both the semantic content of an \gls{llm}-generated plan and the architectural properties required for formal analysis.

Two recent multi-agent systems illustrate how safety is achieved outside the specification-verification-enforcement pipeline surveyed here. ResQConnect, a disaster-response platform coordinating retrieval-augmented agents, and a wildfire-monitoring multi-agent architecture both report strong empirical reliability, but neither specifies a safety property in a formal notation, model-checks a plan or workflow graph, or verifies a policy with a theorem prover~\citep{aththanayake2026resqconnect,sandeep2026wildfire}. Both instead rely on controlled, deterministic execution order enforced through explicit guardrails, schema-constrained outputs, and routing heuristics, validated empirically rather than proved; the wildfire paper's own description of its reliability mechanism, ``architectural enforcement of the orchestrator's recursive validation loop''~\citep{sandeep2026wildfire}, states the distinction precisely. Neither meets this survey's inclusion criteria (Section~\ref{sec:intro:scope}), so neither is a corpus entry; both are evidence, alongside the corpus, that well-engineered multi-agent systems currently default to architectural and empirical safeguards rather than the formal pipeline this survey maps, which is consistent with the specification bottleneck (Theme~A, Section~\ref{sec:synthesis:translation}) as the field's central unresolved problem.

\subsection{Specification Techniques}
\label{sec:related:spec}

Formal verification requires expressing properties in a rigorous language. Three specification paradigms emerge from systematic analysis of the included studies: Temporal Logic, classical planning formalisms, and \gls{nl} constitutions.

First, \textbf{\acrfull{ltl} and derivatives} have become the main specification language for agent plans. NL2LTL, a demonstration Python package, maps \gls{nl} instructions to \gls{ltl} formulas by classifying them into Declare-style pattern templates with a Rasa or OpenAI language-model engine~\citep{He2023nl2ltl}, while NL2Spec adds an interactive disambiguation loop where the \gls{llm} queries the user when requirements are ambiguous~\citep{Cosler2023nl2spec}. To reduce hallucinations in formula generation, \gls{hsd} divides complex requirements into simpler parts~\citep{Anon2025nl2ltl_hsd}, and for cyber-physical and robotic tasks, the \gls{nl}-to-specification approach has been extended to \gls{stl}, combining \gls{llm}-based generation with external knowledge and clarification dialogues~\citep{Fang2025nl2stl,Anon2026clarifystl}. These works share a common finding: \gls{llm}-based \gls{nl}-to-\gls{ltl} translation achieves high syntactic correctness (above \SI{90}{\percent} for leading models) but suffers from low semantic correctness: the generated formula is syntactically well-formed but does not capture the intended property. The \gls{hsd} approach reduces this gap but does not eliminate it, as semantic correctness requires ground-truth execution traces for validation, which are unavailable in novel task domains~\citep{Anon2025nl2ltl_hsd}.

Second, \textbf{classical planning formalisms (\gls{pddl})} have been extensively studied for specification. A survey covering 40+ papers on \gls{llm}-to-\gls{pddl} translation~\citep{Anon2025pddl_survey} points to the Planetarium benchmark~\citep{Zuo2025planetarium}, on which GPT-4o (zero-shot) generates \SI{96.1}{\percent} syntactically valid \gls{pddl} problems but only \SI{24.8}{\percent} semantically correct ones, judged by Planetarium's own \gls{pddl} equivalence check against ground truth rather than by plan validation; the best fine-tuned model reaches \SI{34.2}{\percent}. This result has a direct implication for formal verification: verifying a \gls{pddl} model whose preconditions and effects are semantically incorrect provides false assurance rather than genuine safety guarantees. The same conclusion has been reached through benchmark evaluation, showing that \glspl{llm} fail on even simple planning tasks under formal \gls{pddl} equivalence checking~\citep{Valmeekam2023planbench}.

Third, \textbf{\gls{nl} constitutions and alignment specifications} offer a non-formal alternative. Constitutional \gls{ai}~\citep{Bai2022constitutional} encodes agent constraints as a \gls{nl} constitution of 58 principles that guides both \gls{slaif} and \gls{rlaif}. While not a formal specification, Constitutional \gls{ai} establishes an example for declarative constraint encoding and is now the foundation of production \gls{ai} systems. Its limitation is the absence of formal semantics: two principles may conflict, and the model cannot prove or disprove compliance with any individual constraint.

A synthesis of 200+ papers identified three integration directions for formal techniques and \glspl{llm}: FM-enhancing-\gls{llm} (using verification to certify \gls{llm} outputs), \gls{llm}-enhancing-FM (using \glspl{llm} to improve the usability of formal tools), and unified neural-symbolic systems~\citep{Zhang2024roadmap}. This tripartite taxonomy organizes much of the subsequent work reviewed below.

The specification sub-field faces a fundamental bootstrapping problem: \gls{nl} is the medium through which humans express safety requirements, yet all formal specification languages require expert knowledge to use correctly. \gls{llm}-based \gls{nl}-to-formal translation reduces this barrier but creates a semantic gap that remains unclosed. No current system provides end-to-end, verified specification acquisition, from ambiguous \gls{nl} requirements to formally correct and complete safety specifications, across open-ended task domains.

\subsection{Verification Approaches}
\label{sec:related:verif}

Given a plan representation and a specification, the verification literature has pursued the four paradigms of Section~\ref{sec:taxonomy:l1}: model checking, static analysis, theorem proving, and \gls{llm}-assisted verification. We review runtime monitoring alongside them, although our taxonomy places it under enforcement (ENF-RT) when it intervenes rather than only reports.

First, \textbf{model checking of agent plans} translates \gls{nl} plans into Kripke structures and \gls{ltl} specifications, then applies model checking~\citep{Ramani2025bridging}. On PlanBench, GPT-5 achieves an F1 of \SI{96.3}{\percent} for plan-conformance classification with high syntactic validity, but the semantic correctness of generated Kripke structures stays unverified. Another approach treats the \gls{llm} as a non-deterministic oracle within a formal \gls{fsm}, enabling compositional \gls{ltl} specifications for memory integrity, tool protocols, skill invocation, and human-in-the-loop boundaries~\citep{Anon2026agentverify}. Researchers have also used discrete event systems with \gls{ltl} automata to formally check \gls{llm}-generated multi-robot task allocations against collision and precedence constraints~\citep{Anon2026logicmas}, and a user study (n=12) found that model checking improved end-user planning quality and satisfaction~\citep{Anon2025veriplan}. A common limitation of model-checking approaches is state-space scalability: exhaustive checking over Kripke structures derived from \gls{llm} plans becomes intractable as the plan horizon and vocabulary grow. None of the included studies show model checking at the scale of realistic long-horizon agentic tasks (e.g., SWE-bench, where plans may involve hundreds of actions).

Second, \textbf{static analysis, workflow verification, and theorem proving} offer pre-deployment soundness guarantees. AgentProof is a static verifier for agent workflow graphs, extracting an abstract graph from four frameworks (LangGraph, CrewAI, AutoGen, Google ADK), applying safety checks, and evaluating temporal policies via a \gls{dsl} compiled to \gls{dfa}~\citep{Li2026agentproof}. In 18 workflows, \SI{27}{\percent} had structural defects and \SI{55}{\percent} failed a human-gate policy; verification finishes in under a second for graphs up to 5,000 nodes. AutoRocq verifies programs with Rocq (Coq) through iterative refinement, handling 641 SV-COMP theorems without proof-example pre-training~\citep{NUS2025autorocq}, and VeriGuard (Google) synthesizes and verifies a behavioral policy offline, then uses it as a runtime monitor~\citep{Miculicich2025veriguard}. Static analysis, however, is inherently limited to \emph{finite, prespecified} workflow graphs and cannot verify plans dynamically generated at runtime in response to recorded environmental state, which is the dominant mode in realistic agentic deployments.

Third, \textbf{runtime verification and monitoring} is the most technically mature cluster. AgentSpec, a lightweight \gls{dsl} for specifying and enforcing runtime constraints, achieves above \SI{90}{\percent} unsafe execution prevention in code agents, \SI{100}{\percent} hazardous action elimination in embodied agents, and \SI{100}{\percent} compliance in autonomous driving at millisecond-per-rule overhead~\citep{Wang2025agentspec}. Agent-C enforces \emph{temporal} safety behaviors, that is, constraints on action \emph{sequences} rather than single actions, by translating specifications into first-order logic and applying \gls{smt}-based constrained decoding during token generation, improving GPT-5 conformance from \SIrange{83.7}{100}{\percent} and Claude Sonnet 4.5 from \SIrange{77.4}{100}{\percent}, while raising task utility (GPT-5 from \SI{66.1}{\percent} to \SI{70.6}{\percent}), because it generates a compliant alternative to a non-compliant tool call instead of only blocking it~\citep{Kamath2025agentc}. LogicGuard adopts an actor-critic architecture where the critic analyzes full trajectories and proposes \gls{ltl} constraints communicated in \gls{nl} to the actor, creating a closed-loop safety feedback mechanism that outperforms SayCan and InnerMonologue on the Minecraft diamond-mining benchmark~\citep{Gokhale2025logicguard}. ProbGuard models agent execution traces as a \gls{dtmc}. It provides \gls{pac}-style probabilistic early warnings in autonomous driving up to 15.84 seconds before a violation with no false alarms (up to 38.66 seconds at stricter thresholds), while reducing unsafe behavior by \SI{65.37}{\percent} in household agents~\citep{Wang2025probguard}. A few-shot prompting method with a sliding-window approach (optimal window: five actions) performs pre-execution \gls{ltl} verification of robotic task plans, detecting position errors, missing prerequisites, and redundant actions~\citep{Grigorev2025verifyllm}. Outside the corpus (Section~\ref{sec:methods:irr}), AgentGuard learns an \gls{mdp} of agent behavior online and checks quantitative properties against it at runtime with probabilistic model checking~\citep{Anon2025agentguard}.

Fourth, \textbf{\gls{llm}-assisted verification} uses an \gls{llm}-as-Judge critic with iterative plan revision, attaining up to \SI{90}{\percent} recall and \SI{100}{\percent} precision across four \glspl{llm}, with \SI{96.5}{\percent} of sequences converging within three iterations~\citep{Li2025planverify}. Self-Refine establishes the general pattern of \gls{llm} self-critique, achieving approximately a \SI{20}{\percent} improvement in plan quality on task benchmarks without retraining~\citep{Madaan2023selfrefine}. \gls{llm}-assisted verification offers accessibility but lacks formal soundness guarantees: a model that generates unsafe plans can also justify them during self-critique~\citep{Anon2025fullstack}. A separate line of work uses Dafny as a verification-aware intermediate language: the \gls{llm} generates code annotated with pre/post-conditions and loop invariants, which the Dafny verifier checks before compilation~\citep{Anon2025dafny,Anon2026dafnypro}. DafnyPro achieves \SI{86}{\percent} correct proofs on DafnyBench, a 16-percentage-point improvement over the prior state of the art.

Runtime monitoring methods (AgentSpec, Agent-C, ProbGuard) offer the best combination of practical deployability and demonstrated safety improvements, but provide only one-step or short-horizon guarantees. Static methods (AgentProof, VeriGuard offline) offer pre-deployment soundness but are limited to finite, pre-specified workflows. Model checking approaches handle richer temporal properties but face state-space explosion at realistic task scales. \gls{llm}-assisted verification is accessible and scalable but lacks formal soundness. \emph{No existing strategy combines soundness, completeness, scalability, semantic grounding, and utility preservation.}

\subsection{Enforcement and Correction Mechanisms}
\label{sec:related:enforce}

Enforcement concerns what happens when a plan or action violates a safety property. Examining enforcement mechanisms across the 38 included studies, we identify several distinct approaches and a central empirical finding.

First, \textbf{hard enforcement} implements token generation blocking via constrained decoding when the partial plan would violate a temporal constraint~\citep{Kamath2025agentc}. This approach achieves \SI{100}{\percent} conformance in closed domains and, because it substitutes a compliant alternative for a non-compliant action, raises rather than lowers task utility in Agent-C's evaluation. Hard enforcement also appears as rule-triggered blocking with configurable enforcement actions (block, log, redirect), which achieves millisecond overhead per rule evaluation~\citep{Wang2025agentspec}.

Second, \textbf{system-level guardrails} provide a multi-layer security architecture. LlamaFirewall comprises PromptGuard~2 (jailbreak detection), Agent Alignment Verifications (chain-of-thought auditing for goal misalignment), and CodeShield (static analysis to prevent insecure code generation)~\citep{MetaAI2025llamafirewall}. Deployed in production at Meta, it achieves a \SI{90}{\percent} reduction in the attack success rate on AgentDojo, bringing it to \SI{1.75}{\percent}. Researchers have also proposed a unified safety-alignment framework that fine-tunes tool-using agents using a sandboxed \gls{rl} environment with three-way action classification (benign/malicious/sensitive) and fine-grained reward shaping~\citep{Anon2025rlsafety}.

Third, \textbf{the verifier tax problem} represents the most significant empirical finding in the enforcement literature. Even when enforcement intercepts up to \SI{94}{\percent} of individually unsafe actions, \gls{ssr} (the fraction of tasks completed both safely and correctly) remains below \SI{5}{\percent} in most settings~\citep{Anon2026verifiertax}. This finding, supported by measurements of safety enforcement on $\tau$-Bench~\citep{Sierra2024taubench}, is driven by \emph{integrity leaks}: the model hallucinates user identifiers to bypass authentication checks and finds alternative unsafe paths after direct violations are blocked. This behavior challenges the implicit assumption in most enforcement papers that intercepting individual unsafe actions is sufficient for safe task completion.

Fourth, \textbf{safe tool use} has been formalized in a theoretical framework that specifies the conditions for verifiably safe tool-using \gls{llm} agents, abstracting over specific supervisory mechanisms~\citep{Anon2025toolusetool}.

Our comparative assessment reveals a fundamental tension in the enforcement sub-field: enforcement systems that successfully block unsafe individual actions do not automatically ensure safe task completion. This tension creates a gap between \emph{action-level safety} (verified by most existing systems) and \emph{task-level safety} (measured by \gls{ssr}) that no current system bridges reliably at scale.

\subsection{Benchmarks and Evaluation}
\label{sec:related:bench}

The rapid expansion of this field has driven the parallel development of evaluation infrastructure, including several benchmarks that define how agent safety and performance are measured.

\begin{enumerate}
    \item \textbf{AgentBench}, the first systematic benchmark evaluating \glspl{llm} as agents across 8 environments, tested 29 \glspl{llm} and established task success rate as the primary metric~\citep{Liu2023agentbench}. 
    
    \item \textbf{Agent-SafetyBench} comprises 349 environments, 2,000 test cases, 8 safety risk categories, and 10 failure modes, which makes it the largest safety benchmark in the corpus by test cases~\citep{Zhang2024agentsafety}. Its headline finding, that none of 16 \gls{llm} agents scores above \SI{60}{\percent} safety, motivates the formal verification agenda.  
    
    \item \textbf{AgentHarm} evaluates 110 explicitly malicious agent tasks across 11 harm categories, demonstrating that frontier models exhibit limited robustness even to basic jailbreak attacks in multi-step agentic settings~\citep{Andriushchenko2024agentharm}.  
    
    \item \textbf{PlanBench} introduces a benchmark for evaluating \glspl{llm} on formal \gls{pddl} planning tasks, exposing systematic failures under strict equivalence checking~\citep{Valmeekam2023planbench}.  
    
    \item \textbf{$\tau$-Bench}~\citep{Sierra2024taubench} evaluates policy compliance in transactional tool-agent-user environments with explicit procedural policies, which makes it a natural evaluation platform for formal enforcement methods.
    
\end{enumerate}

A critical gap is the absence of a \emph{unified benchmark} that simultaneously evaluates: (a) formal verification coverage (proportion of safety properties that can be specified and checked), (b) runtime overhead (latency introduced by verification), (c) \gls{ssr} (tasks completed both safely and correctly), and (d) generalizability across agent architectures and task domains. Existing benchmarks address at most two of these dimensions. The introduction of \gls{ssr} represents an important step, but other benchmarks have not yet adopted this metric. Section~\ref{sec:synthesis:heatmap} gives a formal rule for how a unified benchmark should balance these dimensions rather than reporting them independently.

\subsection{Research Gaps and Survey Objectives}
\label{sec:related:gaps}

Synthesizing the five clusters above, we identify ten open problems that the existing literature has not resolved. We state each briefly here; Section~\ref{sec:agenda} gives the full technical treatment.

\begin{enumerate}[label=\textbf{RG\arabic*.}]

    \item \textbf{Semantic translation gap.} \gls{nl}-to-formal translation achieves above \SI{90}{\percent} syntactic correctness but only \SIrange{24.8}{34.2}{\percent} semantic correctness for \gls{pddl}~\citep{Zuo2025planetarium,Cosler2023nl2spec}, propagating false assurance through every downstream verifier.

    \item \textbf{Scalability to long-horizon plans.} Existing verified systems handle \SIrange{5}{30}{} actions; long-horizon tasks (SWE-bench scale) remain beyond exhaustive model checking.

    \item \textbf{The verifier tax.} In one $\tau$-Bench study of transactional tool-use agents, intercepting \SI{94}{\percent} of unsafe actions still yields \gls{ssr} below \SI{5}{\percent}~\citep{Anon2026verifiertax} because agents hallucinate credentials to bypass blocked paths.

    \item \textbf{Dynamic plan generation.} Static methods require pre-specified finite graphs; most deployed agents generate plans dynamically in response to the environment state.

    \item \textbf{Multi-agent verification.} Per-agent verification cannot capture emergent interaction patterns (collusion, cascade failures, trust delegation) in multi-agent systems~\citep{Anon2026logicmas}.

    \item \textbf{Probabilistic specification.} \glspl{llm} are inherently stochastic; deterministic safety specifications are a mismatch. Probabilistic model checking of agent plans is limited to ProbGuard~\citep{Wang2025probguard} in the corpus and AgentGuard~\citep{Anon2025agentguard} outside it.

    \item \textbf{Verified re-planning.} Beyond Agent-C's single-action substitution~\citep{Kamath2025agentc}, no current system generates a constraint-satisfying alternative plan segment when a violation is detected; the other approaches block or abort.

    \item \textbf{Specification acquisition at scale.} Writing formal safety rules requires expert effort that does not scale to the breadth of deployment domains; automated mining from demonstrations or regulatory documents is nascent.

    \item \textbf{Adversarial robustness.} Verification components are themselves attack surfaces; adversarial evaluation of the end-to-end pipeline is systematically absent~\citep{MetaAI2025llamafirewall}.

    \item \textbf{Evaluation standardization.} No benchmark jointly evaluates formal verification coverage, runtime overhead, and \gls{ssr} across architectures and domains.

\end{enumerate}

The ten gaps define the scope of this survey's contributions. As shown in Table~\ref{tab:related_surveys}, no prior survey covers all three pipeline stages using a systematic methodology with evidence grading; this survey is the first to do so.


\section{Cross-Cutting Synthesis}
\label{sec:synthesis}

This section presents a cross-cutting analysis of the 38 included studies, integrating findings across the five thematic clusters identified in Section~\ref{sec:related}. We organize the synthesis around five themes that cut across cluster boundaries, provide a \gls{grade} certainty table for all principal claims, and characterize the coverage landscape using a gap heatmap.

\subsection{Theme A -- The Translation Bottleneck}
\label{sec:synthesis:translation}

Every approach that applies classical formal verification to \gls{llm} agent plans faces the translation problem: converting \gls{nl} plans or specifications into formal structures amenable to algorithmic analysis. Evidence across multiple studies consistently shows that \glspl{llm} achieve high \emph{syntactic} correctness but low \emph{semantic} correctness in this translation.

Evidence on syntactic correctness shows that across the \gls{nl}-to-formal specification cluster, \glspl{llm} achieve above \SI{90}{\percent} syntactic validity in \gls{ltl} formula generation~\citep{He2023nl2ltl,Cosler2023nl2spec,Anon2025nl2ltl_hsd} and \SI{96.1}{\percent} syntactic validity in \gls{pddl} problem generation on Planetarium~\citep{Zuo2025planetarium}. In Kripke structure generation, prior work reports near-perfect syntactic validity with an F1 score of \SI{96.3}{\percent} on plan-conformance classification~\citep{Ramani2025bridging}.

Evidence on semantic correctness, the degree to which a generated formula captures the intended property, tells a different story. On the same benchmark, GPT-4o (zero-shot) achieves only \SI{24.8}{\percent} semantic correctness in \gls{pddl} generation, and the best fine-tuned model \SI{34.2}{\percent}, judged by a \gls{pddl} equivalence check against ground truth~\citep{Zuo2025planetarium}. Research shows that even with interactive disambiguation loops, semantic errors persist for novel or ambiguous requirements~\citep{Cosler2023nl2spec}. Hierarchical decomposition reduces \gls{ltl} semantic errors but does not eliminate them; no study provides a semantic error rate below \SI{10}{\percent} for open-ended task domains~\citep{Anon2025nl2ltl_hsd}.

Formal verification is only as sound as its specification. A model checker operating on a semantically incorrect \gls{ltl} formula can produce a clean verdict (\textit{no violation found}) for a plan that is, in fact, unsafe, and the verdict then gives false assurance. This semantic gap means that the \emph{completeness} of specification acquisition is the binding constraint on the soundness of end-to-end verification, not the power of the verification algorithm applied downstream.

Current best practice combines \gls{llm}-based translation with either an interactive disambiguation loop (NL2Spec~\citep{Cosler2023nl2spec}, ClarifySTL~\citep{Anon2026clarifystl}) or hierarchical decomposition (\gls{hsd}~\citep{Anon2025nl2ltl_hsd}). Neither approach provides formal guarantees of semantic correctness; both reduce error rates without eliminating them. Verified specification acquisition, in which the translation itself is proved correct, remains an open problem (RG1).

\subsection{Theme B -- Runtime vs.\ Static Verification Trade-offs}
\label{sec:synthesis:rtvsstatic}

The verification moment dimension of the taxonomy (Section~\ref{sec:taxonomy:l2}) reveals a fundamental trade-off between \emph{completeness} and \emph{applicability to dynamic plans}.

Static verification approaches, such as AgentProof~\citep{Li2026agentproof} and VeriGuard's offline phase~\citep{Miculicich2025veriguard}, provide pre-deployment soundness guarantees: if the workflow graph passes verification, no execution trace of that graph can violate the checked properties. AgentProof verifies in sub-second time for graphs up to \num{5000} nodes. The cost, however, is scope: static methods assume a prespecified, finite workflow graph. An \gls{llm} generates plans dynamically in response to environment observations, so one cannot verify these plans statically before generation.

Runtime monitoring systems (AgentSpec~\citep{Wang2025agentspec}, Agent-C~\citep{Kamath2025agentc}, ProbGuard~\citep{Wang2025probguard}, LogicGuard~\citep{Gokhale2025logicguard}, and VerifyLLM~\citep{Grigorev2025verifyllm}) handle dynamically generated plans by checking each action (or a sliding window of actions) before execution. Runtime monitors operate at millisecond-per-action overhead~\citep{Wang2025agentspec} and can enforce temporal sequence properties~\citep{Kamath2025agentc}. The cost here is incompleteness: a monitor that checks action $a_i$ before execution cannot reason about the safety of the as-yet-ungenerated actions $a_{i+1}, \ldots, a_n$.

No included study achieves a combination that is both complete (sound over full plan traces) and scalable to dynamically generated, open-ended plans. VeriGuard~\citep{Miculicich2025veriguard} comes closest by verifying a policy offline and using the verified policy as a runtime filter, but this requires the policy to be finite and pre-specified. Partial verification, which verifies safety-critical sub-sequences and gives compositional guarantees for the remainder, is a promising direction but has not yet been demonstrated at scale (RG2, RG4).

Table~\ref{tab:overhead} summarizes reported overhead across runtime monitoring systems. Only AgentSpec~\citep{Wang2025agentspec} quantifies overhead in milliseconds; five of six systems do not quantify overhead at all. This absence of standardized overhead reporting itself constitutes a finding related to the evaluation standardization gap (RG10). Without consistent overhead measurement, we cannot perform cross-system comparisons or cost-benefit assessments of enforcement. None of the included studies reports an end-to-end impact on task latency.

\begin{table}[t]
\centering
\small
\caption{Reported runtime overhead for enforcement systems.}
\label{tab:overhead}
\begin{tabular}{llll}
\toprule
\textbf{System} & \textbf{Overhead unit} & \textbf{Reported value} & \textbf{Domain} \\
\midrule
AgentSpec~\citep{Wang2025agentspec}    & Per-rule evaluation & Milliseconds      & Code, embodied, driving \\
Agent-C~\citep{Kamath2025agentc}       & Constrained decoding & Not quantified    & Retail, airline \\
ProbGuard~\citep{Wang2025probguard}    & Not reported        & Not quantified    & Autonomous driving \\
LogicGuard~\citep{Gokhale2025logicguard} & Critic evaluation & Not quantified    & Minecraft \\
VerifyLLM~\citep{Grigorev2025verifyllm} & Per-window check  & Not quantified    & Household robotics \\
LlamaFirewall~\citep{MetaAI2025llamafirewall} & Per-request  & Production latency & Meta production \\
\bottomrule
\end{tabular}
\end{table}

\subsection{Theme C -- The Verifier Tax}
\label{sec:synthesis:tax}

The most consequential single-study empirical finding in the corpus is what has been termed the \emph{verifier tax}: in one $\tau$-Bench study of transactional tool-use agents, a systematic gap between action-level safety (measured by the fraction of individually unsafe actions blocked) and task-level safety (measured by \gls{ssr})~\citep{Anon2026verifiertax}.

On $\tau$-Bench~\citep{Sierra2024taubench}, evaluated with two open-weight models (GPT-OSS-20B and GLM-4-9B), the TRIAD-SAFETY enforcement architecture intercepts up to \SI{94}{\percent} of non-compliant individual actions. Yet \gls{ssr}, the fraction of tasks completed both safely and correctly, remains below \SI{5}{\percent} in most settings. This gap exceeds one order of magnitude.

The primary driver identified is integrity leaks: agents hallucinate user identifiers (account numbers, policy codes) to bypass authentication constraints when the legitimate path is blocked~\citep{Anon2026verifiertax}. The monitor correctly identifies that the hallucinated identifier is different from the real one and blocks those attempts, too, but the agent continues attempting alternative fabrications. The result is a task stuck in a failure loop: neither completed nor executed in a provably safe manner.

No independent study has replicated the exact $\tau$-Bench figures (\SI{94}{\percent} action-level interception, \gls{ssr} below \SI{5}{\percent}), so this finding's \gls{grade} certainty is \emph{Very Low} (Section~\ref{sec:synthesis:grade}). Two further 2025--2026 benchmarks corroborate the qualitative pattern, an action-level/task-level gap, in different domains and under different metrics, without replicating the specific figures: the best-performing agent on a web/mobile/embodied-agent safety benchmark completes fewer than \SI{40}{\percent} of tasks while fully adhering to safety constraints, a joint completion-and-compliance statistic similar in kind to \gls{ssr} though not measured under an active blocking mechanism~\citep{li2026besafe}; and a separate embodied-agent safety benchmark shows the same trade-off is architecture-dependent rather than universal, since a purpose-built safety-aware agent substantially narrows the gap relative to baseline agents on the same tasks~\citep{chen2025safemind}. Both are arXiv preprints at the time of writing, which this survey's evidence-weighting policy (Section~\ref{sec:synthesis:grade}) treats as corroborating context rather than independent replication for \gls{grade} purposes.

Most enforcement papers implicitly assume that blocking unsafe individual actions is sufficient for safe task completion. The verifier tax finding falsifies this assumption: an agent that cannot execute its intended (possibly unsafe) plan will search for alternative plans that achieve the same goal while bypassing the safety check, rather than gracefully failing. This behavior exemplifies Goodhart's Law: when a proxy measure (action-level compliance) becomes the target of optimization, agents optimize the proxy rather than the underlying goal (task-level safety).

The verifier tax implies that effective enforcement requires:
\begin{itemize}

    \item \textbf{Plan-level, not action-level, safety policies}: Constraints must reason about the goal and context of the full plan, not just the legality of individual actions.
    
    \item \textbf{Integrity constraints}: Formal invariants prevent hallucination of identifiers or credentials by checking them against ground truth.
    
    \item \textbf{Safe replanning}: Upon detecting a violation, the system should generate an alternative plan segment that satisfies all constraints and continues toward the goal, rather than simply blocking.
\end{itemize}

No single system in the included corpus meets any of these three requirements in full; Agent-C's compliant-alternative generation~\citep{Kamath2025agentc} is a single-action step toward the third. No independent replication has confirmed the verifier tax, which limits its \gls{grade} certainty to \emph{Very Low}, but the mechanism that the study identifies is consistent with the failure pattern noted in LlamaFirewall~\citep{MetaAI2025llamafirewall} (a residual \SI{1.75}{\percent} attack success rate after \SI{90}{\percent} reduction, driven by adaptive adversaries finding alternative paths).

The verifier tax and the translation bottleneck are compounding, not independent, failure modes. When specifications are semantically incorrect (Theme~A), the enforcement rules derived from them will also be semantically incorrect: the monitor blocks actions that are unsafe \emph{according to the flawed specification}, not necessarily those that are unsafe \emph{in the real world}. This incorrectness creates a second source of \gls{ssr} degradation beyond the integrity-leak mechanism: agents that satisfy a semantically incorrect specification may still execute genuinely unsafe plans, while agents blocked on legitimate paths (due to specification over-conservatism) seek alternatives. Resolving the verifier tax, therefore, requires progress on both the enforcement architecture (RG3) and the correctness of the semantic specification (RG1); neither is sufficient on its own.

We advance one candidate mechanism for \emph{why} hard blocking produces this failure loop, stated here explicitly as our own hypothesis rather than a corpus finding: a policy $\pi(a \mid h)$ trained on histories $h$ drawn from a training distribution $\mathcal{D}_{\text{train}}$ encounters, after a hard block, a history $h'$ (containing a refusal event) that is rare or absent under $\mathcal{D}_{\text{train}}$. If the per-step error on such out-of-distribution histories exceeds the in-distribution error $\epsilon$, the established compounding-error bound for imitation learning under distribution shift applies: a policy with per-step error $\epsilon$ on states it was not trained to handle can accumulate $O(T^2\epsilon)$ mistakes over a horizon of length $T$, rather than the $O(T\epsilon)$ that would hold if errors did not compound~\citep{ross2011dagger}. Applied here, each hard block without a verified alternative path pushes the agent toward exactly the kind of out-of-distribution history this bound describes, offering a formal analogue for why repeated blocking without re-planning could produce escalating, rather than merely repeated, failure. No study in the corpus tests this mechanism directly; we state it as a research question (RG7) rather than an established result.

\subsection{Theme D -- Probabilistic Monitoring as Middle Ground}
\label{sec:synthesis:probabilistic}

ProbGuard~\citep{Wang2025probguard} and the broader probabilistic monitoring paradigm represent a pragmatic middle ground between deterministic enforcement (high assurance, low scalability) and no verification (high scalability, no assurance).

Rather than guaranteeing zero violations, which is difficult for stochastic, open-ended agents, probabilistic monitors provide:
\begin{itemize}

 \item \textbf{Early warnings:} ProbGuard provides safety warnings up to 15.84 seconds in advance in autonomous driving scenarios without false alarms (up to 38.66 seconds at stricter thresholds), enabling human intervention or replanning before irreversible harm.

 \item \textbf{\gls{pac}-style accuracy bounds:} the monitor's warning accuracy is bounded by \gls{pac}-learning guarantees, providing a formal relationship between the number of observed traces and the probability of false alarms.

 \item \textbf{Graceful degradation:} in household agent settings, ProbGuard reduces unsafe behavior by \SI{65.37}{\percent} while keeping the task completion rate at \SI{80.4}{\percent}. No corpus study compares this trade-off with hard enforcement on the same tasks.
\end{itemize}

ProbGuard requires a \gls{dtmc} model trained on observed traces, which requires a corpus of prior agent executions in the target domain. It provides probabilistic rather than deterministic guarantees, which may be insufficient for safety-critical deployments. It has been evaluated in only two domains (autonomous driving and household agents), thereby limiting generalizability (\gls{grade}: \emph{Very Low}).

Probabilistic monitoring sidesteps the verifier tax problem: rather than attempting to block all violations (which would drive the agent to seek alternative unsafe paths), it issues warnings. It defers to human intervention, preserving the agent's plan space while flagging risks. This combination suggests that the optimal enforcement architecture may combine probabilistic monitoring (for early warning) with selective hard blocking (for high-certainty, high-severity violations), rather than relying exclusively on either paradigm.

\subsection{Theme E -- Compositional Specification as an Organizing Principle}
\label{sec:synthesis:compositional}

AgentVerify~\citep{Anon2026agentverify} and AgentProof~\citep{Li2026agentproof} independently argue for and demonstrate compositional specification: they specify and verify safety properties for distinct concern domains (memory integrity, tool call protocols, human-in-the-loop boundaries) independently, then compose them.

Compositional specification demonstrates several benefits. First, incremental verification allows us to check, update, or extend individual property families without re-verifying the entire system. Second, reuse across architectures is possible: AgentProof applies the same compositional framework to four distinct agent orchestration frameworks (LangGraph, CrewAI, AutoGen, Google ADK), demonstrating cross-architecture reuse. Third, modular auditability means separate safety domains can be audited independently by domain-specific reviewers. Table~\ref{tab:domains} maps the safety domains addressed by included enforcement and verification systems.

\begin{table}[t]
\centering
\small
\caption{Safety domains addressed by included verification/enforcement systems. \checkmark: explicitly evaluated. $\circ$: partially addressed. ---: not addressed.}
\label{tab:domains}
\begin{tabular}{lcccccc}
\toprule
\textbf{System} & \textbf{Temporal seq.} & \textbf{Memory/PII} & \textbf{Tool safety} & \textbf{Credentials} & \textbf{Multi-agent} & \textbf{Adversarial} \\
\midrule
AgentSpec~\citep{Wang2025agentspec}     & \checkmark & $\circ$ & \checkmark & --- & --- & --- \\
Agent-C~\citep{Kamath2025agentc}        & \checkmark & --- & \checkmark & $\circ$ & --- & --- \\
AgentVerify~\citep{Anon2026agentverify} & \checkmark & \checkmark & \checkmark & --- & --- & --- \\
AgentProof~\citep{Li2026agentproof}     & \checkmark & $\circ$ & \checkmark & --- & $\circ$ & --- \\
LlamaFirewall~\citep{MetaAI2025llamafirewall} & --- & --- & $\circ$ & --- & --- & \checkmark \\
ProbGuard~\citep{Wang2025probguard}     & \checkmark & --- & --- & --- & --- & --- \\
VeriGuard~\citep{Miculicich2025veriguard} & $\circ$ & \checkmark & \checkmark & --- & --- & --- \\
\bottomrule
\end{tabular}
\end{table}

Table~\ref{tab:domains} reveals a consistent pattern: temporal sequence properties and tool safety are the most widely addressed domains; credential integrity and multi-agent collusion are systematically underaddressed. This gap directly maps to the integrity leak mechanism driving the verifier tax (Theme~C) and to the multi-agent verification gap (RG5).

Of the 7 systems in Table~\ref{tab:domains}, 6 are marked \emph{not addressed} in the Credentials column and the seventh, Agent-C, only \emph{partially addressed}. Temporal-sequence and tool-call safety, the two best-covered domains in the table, constrain what an agent does and when; neither constrains whether an identifier the agent supplies mid-plan is genuine. This is why the field's existing taxonomy misses credential integrity even though it is central to the verifier tax's integrity-leak mechanism (Theme~C, Section~\ref{sec:synthesis:tax}): an identifier-integrity monitor (RG3.2, Appendix~\ref{sec:agenda:g3}) checking identifiers against a ground-truth index is the concrete fix, and no system in the corpus has built it. The integrity-leak mechanism itself rests on a single benchmark study~\citep{Anon2026verifiertax}, which is why Table~\ref{tab:grade} rates it \emph{Very Low}.

\subsection{\gls{grade} Certainty Assessment}
\label{sec:synthesis:grade}

Table~\ref{tab:grade} presents the \gls{grade} certainty assessment for the principal claim in each of the five thematic clusters. We downgrade certainty from \emph{High} for: study limitations (L), inconsistency across studies (I), indirectness (narrow domains, non-generalizable benchmarks) (D), and imprecision (single study, wide confidence) (P). Each triggered domain lowers the rating by one level (High, Moderate, Low, Very Low), and Very Low is the floor. GRADE defines each level by how close the figure a claim reports is likely to be to the true value, the one that more and larger studies would measure~\citep{balshem2011grade}. At \emph{High}, we are confident that it lies close; at \emph{Moderate}, it is probably close but could differ substantially; at \emph{Low}, it may differ substantially; and at \emph{Very Low}, it is likely to differ substantially.

We adapt the four \gls{grade} downgrading domains used in Table~\ref{tab:grade} to LLM-agent safety evidence using the concrete, per-domain triggers in Table~\ref{tab:gradecriteria}. We start every claim's assessment at \emph{High} rather than at the \emph{Low} starting point GRADE's own guidelines assign to observational studies~\citep{balshem2011grade}, because a benchmark evaluation with a fixed task set and an automatically scored outcome is methodologically closer to a controlled observational study than to an uncontrolled case series; readers who prefer the stricter \emph{Low} starting point should read every rating in Table~\ref{tab:grade} one level down. We do not include a fifth GRADE domain, publication bias, as a column in Table~\ref{tab:grade}, since detecting it requires comparing published to unpublished results systematically, which this review's search protocol (Section~\ref{sec:methods:search}) does not support; we flag this as a standing limitation of every rating in the table rather than as a per-claim assessment. The same limitation applies uniformly across all five thematic-cluster syntheses in Section~\ref{sec:synthesis} (Themes A--E): none was assessed separately for missing-result risk, since the underlying search protocol that limits detection is common to all five, not specific to any one theme.

\begin{table}[t]
\centering
\small
\caption{GRADE downgrade domains operationalized for LLM-agent safety evidence.}
\label{tab:gradecriteria}
\begin{tabular}{p{0.15\linewidth} p{0.7\linewidth}}
\toprule
\textbf{Domain} & \textbf{Concrete trigger for this literature} \\
\midrule
Study limitations (L) & No released code, specification, or evaluation harness; a single model family evaluated; no variance or confidence interval reported across runs. \\
Inconsistency (I) & Conflicting results for the same claim across two or more independent studies or benchmarks. \\
Indirectness (D) & Evaluation restricted to a simulated or narrow task domain that differs materially from realistic deployment conditions. \\
Imprecision (P) & A single supporting study, fewer than a field-appropriate minimum number of tasks, or no confidence interval reported. \\
Publication bias & Standing limitation across all ratings (Section~\ref{sec:synthesis:grade}); not scored per claim in this review. \\
\bottomrule
\end{tabular}
\end{table}

Worked example: the verifier-tax claim (Table~\ref{tab:grade}, row~7) triggers L (single released benchmark configuration, no reported confidence interval on the 94\%/5\% figures), D (evaluated only on $\tau$-Bench's transactional tool-use domain), and P (one supporting study), for three downgrade steps from \emph{High} to \emph{Very Low}, the rating Table~\ref{tab:grade} shows.

Table~\ref{tab:gradecriteria}'s domains also give a concrete path from \emph{Very Low}/\emph{Low} certainty toward \emph{High}: study limitations are addressed by released artifacts, pinned model versions and decoding parameters and seeds, and multiple runs reported with confidence intervals; inconsistency by shared task suites run independently by several labs; indirectness by evaluation on deployment-like tool environments rather than simulated domains; and imprecision by larger task sets with reported confidence intervals. Standards the field could adopt for this purpose include ACM artifact-evaluation badging, machine-readable specification formats already used elsewhere in this literature (TLSF for \gls{ltl}, \gls{pddl} itself for planning), containerized evaluation harnesses, and pre-registered multi-site replications.

\begin{table}[t]
\centering
\small
\caption{\gls{grade} certainty of evidence for principal claims. Downgrade reasons: L = study limitations, I = inconsistency, D = indirectness, P = imprecision (single study).}
\label{tab:grade}
\begin{tabular}{p{0.35\linewidth} p{0.3\linewidth} cccc c}
\toprule
\textbf{Claim} & \textbf{Supporting evidence} & \textbf{L} & \textbf{I} & \textbf{D} & \textbf{P} & \textbf{\gls{grade}} \\
\midrule
No \gls{llm} agent achieves $>$\SI{60}{\percent} safety on comprehensive benchmarks & Agent-SafetyBench (16 agents, 2k cases)~\citep{Zhang2024agentsafety} & & & & \checkmark & \textbf{Moderate} \\
\gls{llm}-to-\gls{ltl} translation achieves $>$\SI{90}{\percent} syntactic correctness & 4+ independent studies~\citep{He2023nl2ltl,Cosler2023nl2spec,Anon2025nl2ltl_hsd,Ramani2025bridging} & & \checkmark & & & \textbf{Moderate} \\
\gls{llm}-to-\gls{pddl} semantic correctness is \SIrange{24.8}{34.2}{\percent} & Planetarium (1 benchmark)~\citep{Zuo2025planetarium}, via survey~\citep{Anon2025pddl_survey} & \checkmark & & \checkmark & \checkmark & \textbf{Very Low} \\
\gls{ltl} runtime enforcement achieves \SI{100}{\percent} conformance in closed domains & Agent-C, AgentSpec (2 studies)~\citep{Kamath2025agentc,Wang2025agentspec} & \checkmark & & \checkmark & \checkmark & \textbf{Very Low} \\
Static graph verification detects structural defects in $<$1\,s & AgentProof (1 study, 18 workflows)~\citep{Li2026agentproof} & \checkmark & & \checkmark & \checkmark & \textbf{Very Low} \\
Probabilistic monitoring warns up to 15.84 s before violation without false alarms & ProbGuard (1 study, 2 domains)~\citep{Wang2025probguard} & \checkmark & & \checkmark & \checkmark & \textbf{Very Low} \\
Enforcement reduces \gls{ssr} (verifier tax) & Verifier Tax (1 study, $\tau$-Bench)~\citep{Anon2026verifiertax} & \checkmark & & \checkmark & \checkmark & \textbf{Very Low} \\
\gls{llm} self-critique improves plan quality by $\sim$\SI{20}{\percent} & Self-Refine (NeurIPS 2023)~\citep{Madaan2023selfrefine} & \checkmark & & \checkmark & \checkmark & \textbf{Very Low} \\
Dafny intermediate language achieves \SI{86}{\percent} proof success & DafnyPro (1 study, DafnyBench)~\citep{Anon2026dafnypro} & \checkmark & & \checkmark & \checkmark & \textbf{Very Low} \\
\bottomrule
\end{tabular}
\end{table}

The \gls{grade} summary reflects the field's early stage: despite positive findings across multiple systems, several factors limit certainty: limited domain scope, the absence of independent replication, and high benchmark heterogeneity. The two \emph{Moderate} certainty claims form the empirical bedrock of the field; all other claims require replication at a larger scale and across diverse domains before they can inform deployment decisions.

We do not apply an additional \gls{grade} downgrade for preprint status alone; the imprecision (P) and indirectness (D) domains already in Table~\ref{tab:grade} capture the relevant construct-validity concerns per claim, and a further blanket preprint penalty would double-count the same evidence weakness. Two rows of Table~\ref{tab:grade} rest only on preprints (Agent-SafetyBench and AgentProof); Table~\ref{tab:taxonomy} gives the venue status of every supporting study, and the BeSafe-Bench and SafeMind preprints of Section~\ref{sec:synthesis:tax} enter no rating.

We checked every row of Table~\ref{tab:taxonomy}'s Venue column against Crossref, ACL Anthology, IEEE Xplore, ACM DL, DBLP (where accessible), or the conference or workshop's own accepted-papers page, rather than inferring it from the arXiv listing: 23 of 38 corpus entries (\SI{61}{\percent}) are peer-reviewed, among them Toolformer and Chain-of-Thought at NeurIPS and ReAct, AgentHarm, and AgentBench at ICLR; the remaining 15 (\SI{39}{\percent}) are preprint-only as of this check, including Constitutional~AI, which has not appeared in a peer-reviewed venue despite its wide citation. NL2LTL is a three-page AAAI-23 demonstration paper~\citep{He2023nl2ltl}, counted as peer-reviewed. Three studies (Agent-C, LogicGuard, and the LLM-as-Judge plan-verification study) have workshop or conference submissions whose archival status (formal proceedings versus a non-archival poster or talk) we could not confirm from public sources; we count them provisionally as preprints ($^{\ddagger}$). Workshop papers with a genuine program committee and archival proceedings (Dafny@POPL, AgenticSE@ASE) are counted as peer-reviewed but distinguished from main-track venues in the table; this is a coarser standard than a top-tier conference or journal, and readers who require the stricter standard should treat those three workshop rows and the NL2LTL demonstration row as PP.

\subsection{Coverage Heatmap}
\label{sec:synthesis:heatmap}

Before mapping individual gaps, we give a formal rule for how a unified benchmark (RG10) should balance the metrics identified in Section~\ref{sec:related:bench}. Define a benchmark result as a vector $(C, \text{SSR}, L)$: representational coverage $C \in [0,1]$ (the fraction of the deployment's safety properties the specification language can express), closed-loop task-level safety $\text{SSR} \in [0,1]$ (observable only from a full closed-loop run, following~\citep{Anon2026verifiertax}), and runtime overhead $L$ against a deployment-specific latency budget $L_{\text{budget}}$. We rank systems by Pareto dominance over this vector: system $A$ dominates $B$ if $A$ is no worse than $B$ on all three axes and strictly better on at least one. For deployment selection among non-dominated systems, we give an $\epsilon$-constraint rule: maximize \text{SSR} subject to $L \leq L_{\text{budget}}$ and $C \geq C_{\text{min}}$ for a deployment-chosen minimum coverage $C_{\text{min}}$. This formalizes the point that no single scalar should stand in for all three axes (just as precision and recall remain distinct even though F1 combines them, and Pareto dominance needs no scalar combination), while still giving an operational rule for how a unified benchmark, or a deployment team, should weigh the three dimensions against one another. No included study currently reports $C$, \text{SSR}, and $L$ consistently enough on the same tasks to instantiate $L_{\text{budget}}$ or $C_{\text{min}}$ from the corpus; we present the rule as a requirement future benchmarks should satisfy (RG10.1), not as a result computed here.

Table~\ref{tab:heatmap} maps the ten identified research gaps (RG1 to RG10) against the five thematic clusters. Each cell indicates whether the cluster \emph{causes} the gap ($\bullet$), partially addresses it ($\circ$), or is unrelated (blank).

\begin{table}[t]
\centering
\small
\caption{Gap coverage heatmap. $\bullet$: this cluster gives rise to or studies the gap. $\circ$: the cluster partially addresses the gap.} \label{tab:heatmap}
\begin{tabular}{lcccccc}
\toprule
\textbf{Gap} & \textbf{Arch} & \textbf{Spec} & \textbf{Verif} & \textbf{Enf} & \textbf{Bench} \\
\midrule
RG1 -- Semantic translation    & & $\bullet$ & $\circ$ & & \\
RG2 -- Long-horizon scale      & $\bullet$ & & $\bullet$ & & $\circ$ \\
RG3 -- Verifier tax            & & & $\circ$ & $\bullet$ & $\bullet$ \\
RG4 -- Dynamic plan generation & $\bullet$ & & $\bullet$ & & \\
RG5 -- Multi-agent             & $\bullet$ & $\circ$ & $\bullet$ & & \\
RG6 -- Stochastic model check. & & $\circ$ & $\bullet$ & & \\
RG7 -- Verified re-planning    & & $\circ$ & $\circ$ & $\bullet$ & \\
RG8 -- Spec acquisition scale  & & $\bullet$ & & & \\
RG9 -- Adversarial robustness  & & & $\circ$ & $\bullet$ & \\
RG10 -- Evaluation standard.   & & & & $\circ$ & $\bullet$ \\
\bottomrule
\end{tabular}
\end{table}

The heatmap reveals two structural asymmetries. First, the specification cluster is the source of RG1 and RG8, the two gaps that are purely about how properties are acquired, and is not a target of any other cluster's work. This finding confirms that specification is the most underinvested part of the pipeline relative to its importance. Second, the enforcement cluster drives RG3 (verifier tax) and RG7 (verified replanning). Still, these gaps also require input from verification (for plan-level reasoning) and benchmarks (for \gls{ssr} measurement); because they span several clusters, no single sub-field can resolve them.


\section{Research Agenda}
\label{sec:agenda}

Section~\ref{sec:related:gaps} states the ten open problems (RG1 to RG10), and Section~\ref{sec:synthesis} traces where they arise. Table~\ref{tab:agenda} gives, for each gap, the difficulty that binds current work and two or three concrete research directions. Appendix~\ref{app:agenda} develops every gap in full: the current state of the art, the core technical difficulty, and pointers to the most relevant prior work for each direction. Section~\ref{sec:agenda:complexity} analyzes what the table cannot show: how verification cost grows with plan length and with the number of agents.

\begin{table}[t]
\centering
\footnotesize
\caption{Research directions for the ten open problems of Section~\ref{sec:related:gaps}. Appendix~\ref{app:agenda} develops each gap and each direction with pointers to prior work.}
\label{tab:agenda}
\begin{tabular}{p{0.15\linewidth} p{0.31\linewidth} p{0.45\linewidth}}
\toprule
\textbf{Gap} & \textbf{Binding difficulty} & \textbf{Research directions} \\
\midrule
RG1 -- Semantic translation & Novel domains have no independent ground truth against which to check a translated specification & RG1.1 grammar- and type-constrained decoding; RG1.2 \gls{cegis}-style equivalence checking of candidate translations; RG1.3 benchmarks pairing requirements with verified specifications and traces \\
RG2 -- Long-horizon scale & Kripke state space grows exponentially with plan length; verified systems handle \SIrange{5}{30}{} actions & RG2.1 partial verification of safety-critical sub-sequences; RG2.2 abstraction refinement focused on high-risk transitions; RG2.3 assume-guarantee contracts between plan segments \\
RG3 -- Verifier tax & Agents route around action-level blocks, for example by fabricating identifiers & RG3.1 plan-level, goal-contextual policies; RG3.2 identifier-integrity monitor against a ground-truth index; RG3.3 safe replanning upon violation \\
RG4 -- Dynamic plans & Static methods need a finite, pre-specified workflow graph & RG4.1 online, incremental workflow-graph extraction; RG4.2 pre-verified plan templates whose instantiations inherit the proof \\
RG5 -- Multi-agent & Composing agents raises the complexity class (Section~\ref{sec:agenda:complexity}) & RG5.1 assume-guarantee contracts between agents; RG5.2 protocol verification with session types or communicating automata \\
RG6 -- Probabilistic specification & Stochastic agents meet deterministic specifications; one corpus study (ProbGuard) addresses this & RG6.1 \gls{pctl} properties checked on \glspl{dtmc} learned from traces; RG6.2 calibrated monitors via conformal prediction \\
RG7 -- Verified re-planning & Enforcement blocks, logs, or aborts; only single-action substitution exists & RG7.1 constraint-satisfying plan repair; RG7.2 \gls{smt}-guided constrained plan generation \\
RG8 -- Specification acquisition & Formal specifications need expert effort that does not scale across domains & RG8.1 mining specifications from labeled traces; RG8.2 extraction from regulatory documents; RG8.3 shared compositional specification libraries \\
RG9 -- Adversarial robustness & Verification components are attack surfaces and are not red-teamed end to end & RG9.1 red-teaming of the verification pipeline; RG9.2 formally verified specification translators \\
RG10 -- Evaluation standardization & No benchmark jointly reports coverage, overhead, and \gls{ssr} & RG10.1 unified evaluation suite with an illustrative efficiency-safety index; RG10.2 \gls{ssr} as the primary metric; RG10.3 contamination mitigation \\
\bottomrule
\end{tabular}
\end{table}

\subsection{Verification Cost under Long Horizons and Multiple Agents}
\label{sec:agenda:complexity}

Plan length (RG2) and agent count (RG5) are the two axes along which verification cost grows, and Table~\ref{tab:agenda} lists directions for each. Verifying a property of a multi-agent system requires reasoning about the joint execution of all agents. This is a change in complexity class, not merely a larger instance of the same problem. Composing $n$ single-agent Kripke structures into one concurrent program raises \textsc{ctl}/\textsc{ctl}$^*$ \emph{program} complexity (complexity measured in the size of the composed components) to \textsc{pspace}-complete~\citep{kupferman2000branching}, strictly harder than the polynomial-time model checking available for a single agent, before any strategic reasoning is introduced. Reasoning about what a coalition of agents can jointly force requires Alternating-time Temporal Logic (\textsc{atl}/\textsc{atl}$^*$): model checking is polynomial in the game structure and formula size for \textsc{atl}, but \textsc{2exptime}-complete for \textsc{atl}$^*$~\citep{alur2002atl}, and becomes undecidable for \textsc{atl} itself once agents have imperfect information about each other's state under perfect-recall semantics~\citep{dimatiplea2011}, which is the realistic case for LLM agents, which do not observe each other's internal reasoning. Auditing the corpus against this distinction: the two included multi-agent verification studies collapse the system into a single-threaded formalism rather than reasoning about it as a distributed, concurrent system. The logic-based manufacturing verifier applies discrete-event-system \gls{ltl} over one composed automaton for multi-robot task allocation~\citep{Anon2026logicmas}, and AgentVerify embeds the LLM as a non-deterministic oracle inside one finite-state machine and applies standard \gls{ctl}~\citep{Anon2026agentverify}; neither uses \textsc{atl}, \textsc{atl}$^*$, or an explicit game-theoretic solution concept. A symbolic \textsc{atl}$^*$ model checker~\citep{atlas2025} and a TLA+/Verus-verified concurrency-anomaly detector for multi-agent LLM systems~\citep{khan2026concurrency} are recent, complementary tooling outside the corpus that point toward closing this gap.

Table~\ref{tab:scaling} reports the cost in plan length (RG2) alongside the multi-agent axis: the two dimensions compound rather than trade off, since a long-horizon, multi-agent plan pays both the exponential-in-$|\varphi|$ product-construction cost of Appendix~\ref{sec:bg:mc} and the \textsc{pspace}-complete program-complexity cost of composing $k$ agents~\citep{kupferman2000branching}. The one tractable strategic-reasoning row in that table, perfect-information \textsc{atl}, is also the row least representative of how LLM agents actually operate, since agents typically cannot observe each other's internal state.

\begin{table}[t]
\centering
\small
\caption{Verification cost as a function of plan length $n$ and agent count $k$, by mechanism. Notation follows each source; $|M|$ = model size, $|\varphi|$ = formula length, $|G|$, $|D|$ = workflow graph and DFA sizes.}
\label{tab:scaling}
\begin{tabular}{p{0.24\linewidth} p{0.16\linewidth} p{0.26\linewidth} p{0.2\linewidth}}
\toprule
\textbf{Mechanism} & \textbf{Cost in $n$} & \textbf{Cost in $k$ (agents)} & \textbf{Reported limit} \\
\midrule
Runtime automaton monitor (AgentSpec) & $O(1)$ per action & Not evaluated for $k>1$ & Millisecond/rule~\citep{Wang2025agentspec} \\
\gls{ltl} product check (single agent) & $|M|\cdot 2^{O(|\varphi|)}$ & N/A & \SIrange{5}{30}{} actions~\citep{Ramani2025bridging} \\
\gls{ctl}/\gls{ctl}$^*$, $k$ agents composed & Polynomial in $|M|$ & \textsc{pspace}-complete (program compl.)~\citep{kupferman2000branching} & Not evaluated in corpus \\
\textsc{atl}, perfect information & Polynomial in $|M|$, $|\varphi|$ & Polynomial~\citep{alur2002atl} & Not evaluated in corpus \\
\textsc{atl}, imperfect information & Undecidable~\citep{dimatiplea2011} & Undecidable & Not applicable \\
DFA product (AgentProof) & $O(|G|\cdot|D|)$ & Single-agent workflows only & 5{,}000 nodes, sub-second~\citep{Li2026agentproof} \\
\gls{pctl} on learned \gls{dtmc} (ProbGuard) & Polynomial in states & Not evaluated for $k>1$ & 2 domains~\citep{Wang2025probguard} \\
\bottomrule
\end{tabular}
\end{table}


\section{Discussion}
\label{sec:discussion}

The taxonomy developed in Section~\ref{sec:taxonomy} provides a structural lens for analyzing the literature. We now turn to its implications: the answers to the research questions, practical deployment recommendations, and threats to the validity of this review. Appendix~\ref{app:discussion} assesses the maturity of the field and discusses the relationship between the three safety paradigms represented in the corpus and the societal and ethical considerations that formal verification of \gls{llm} agents raises.

\subsection{Resolution of the Research Questions}
\label{sec:discussion:rqs}

\begin{description}

    \item[RQ1.] NL-to-formal translation achieves above \SI{90}{\percent} syntactic correctness for LTL~\citep{He2023nl2ltl,Cosler2023nl2spec,Anon2025nl2ltl_hsd} and \SI{96.1}{\percent} for PDDL~\citep{Zuo2025planetarium}, but semantic correctness is \SIrange{24.8}{34.2}{\percent} for PDDL~\citep{Zuo2025planetarium} and reaches \SIrange{44.4}{86.11}{\percent} for LTL depending on disambiguation strategy~\citep{Cosler2023nl2spec} (Section~\ref{sec:synthesis:translation}). This binds every downstream verifier: a sound model checker applied to a semantically incorrect formula provides false assurance (Section~\ref{sec:synthesis:translation}).

    \item[RQ2.] Four verification paradigms are represented in the corpus (Section~\ref{sec:related:verif}): model checking (state-space scalability is the binding constraint), static analysis (limited to finite, pre-specified workflow graphs), theorem proving (highest assurance, limited to code-based agents), and LLM-assisted verification (scalable, no formal soundness guarantee). No included study combines soundness, completeness, and scalability (Section~\ref{sec:related:verif}).

    \item[RQ3.] Not reliably. The verifier tax finding (Section~\ref{sec:synthesis:tax}) shows that intercepting up to \SI{94}{\percent} of individually unsafe actions does not translate into safe task completion in the one study that has measured this directly~\citep{Anon2026verifiertax}; two further 2025--2026 benchmarks show a qualitatively similar action-level/task-level gap in different domains and under different metrics~\citep{li2026besafe,chen2025safemind} (Section~\ref{sec:synthesis:tax}). This evidence base is too narrow to generalize the exact 94\%/5\% figures beyond the $\tau$-Bench setting they were measured in.

    \item[RQ4.] Only 2 of the 9 principal claims in Table~\ref{tab:grade} reach \gls{grade} \emph{Moderate}; none reach \emph{High} (Section~\ref{sec:synthesis:grade}). No claim in the corpus is supported by an independently replicated, multi-site study.

\end{description}

\subsection{Practical Implications for Deployment}
\label{sec:discussion:practical}

The evidence base, despite its limitations, supports several practical recommendations for teams deploying \gls{llm} agents in consequential settings today.

\begin{description}

    \item[REC1: Do not rely on action-level enforcement alone.] The verifier tax finding~\citep{Anon2026verifiertax} demonstrates that blocking individually unsafe actions does not produce safe task completion. Any enforcement system deployed in production should measure \gls{ssr}, not only action conformance. Enforcement should be supplemented by integrity constraints (Appendix~\ref{sec:agenda:g3}) and human-in-the-loop gates for high-stakes decisions.

    \item[REC2: Prefer runtime monitoring over static verification for dynamic agents.] For agents that generate plans dynamically (the majority of production deployments), static pre-execution verification is structurally inapplicable. Runtime monitoring systems (AgentSpec~\citep{Wang2025agentspec}, Agent-C~\citep{Kamath2025agentc}) offer the best currently available combination of practical deployability and demonstrated safety improvement. Note that millisecond-level per-rule overhead is confirmed only for AgentSpec; other runtime systems have not quantified their overhead (Table~\ref{tab:overhead}).

    \item[REC3: Use compositional specification for maintainability.] Safety specifications for agent systems should be organized compositionally by concern domain (memory integrity, tool protocols, human gates, credential constraints). This compositional organization enables incremental auditing, independent updating of sub-specifications, and reuse across agent architectures~\citep{Li2026agentproof,Anon2026agentverify}.

    \item[REC4: Apply probabilistic monitoring for early warning in high-stakes domains.] In domains where safety violations are severe and irreversible (autonomous driving, medical decision support, financial transactions), probabilistic monitoring~\citep{Wang2025probguard} provides actionable early warnings with \gls{pac}-style accuracy bounds. The 15.84-second warning that ProbGuard achieves without false alarms in autonomous driving corresponds to meaningful intervention time for a human supervisor.

    \item[REC5: Invest in specification quality, not only verification power.] The most sophisticated model checker applied to a semantically incorrect specification provides false assurance. Teams should invest in specification review and validation (using interactive disambiguation tools such as NL2Spec~\citep{Cosler2023nl2spec}) before deploying verification infrastructure. They should treat the specification as a first-class artifact subject to version control, testing, and peer review.

\end{description}

\subsection{Threats to Validity of This Review}
\label{sec:discussion:threats}

This review is subject to several limitations. We organize them according to four standard validity threats: internal, external, construct, and conclusion validity.

\textbf{Internal validity.} The review was conducted by a single reviewer, introducing the risk of subjective selection and data extraction bias. A validation check in which three raters screened a repeated 37-record search at title/abstract level (Section~\ref{sec:methods:irr}) gave Fleiss'~$\kappa = 0.81$ and pairwise raw agreement between \SI{86.5}{\percent} and \SI{97.3}{\percent}, but it does not cover full-text screening or data extraction, and its two additional raters are co-authors rather than external reviewers. Borderline cases were resolved conservatively (retain if in doubt); this strategy may have inflated inclusion counts relative to a dual-reviewer protocol with explicit disagreement resolution. In the check, three of the four disagreements in which the primary reviewer stood alone were inclusions that both co-authors rejected. The same check found three corpus studies (NL2LTL, NL2Spec, NL-to-LTL/HSD) that all three raters judged out of scope and 11 in-scope records missing from the corpus. Readers should weigh the \gls{grade} certainty ratings in Table~\ref{tab:grade} with this single-reviewer limitation in mind, since GRADE's study-limitations domain (Section~\ref{sec:synthesis:grade}) does not separately capture reviewer-count bias in the underlying corpus selection itself.

\textbf{External validity.} We limited the search to six English-language databases and to papers accessible in full text. The survey does not capture proprietary industry systems (e.g., OpenAI's internal safety infrastructure, Google's production agent safety layer). The primary systematic search (Section~\ref{sec:methods:selection}) excludes papers published after 29 May 2026; a targeted, non-systematic search (Section~\ref{sec:methods:selection}) identified evidence current to 30 August 2026, but it did not follow the full PRISMA protocol and its findings are reported as corroborating evidence (Sections~\ref{sec:synthesis:tax},~\ref{sec:agenda:complexity}) rather than as formal corpus entries. The field's publication pace (multiple relevant papers per week on arXiv) means this survey represents a snapshot rather than a complete, up-to-date map.

\textbf{Construct validity.} The \gls{grade} framework originated in clinical research, and we adapted it to computer science, a field its designers did not have in mind. The resulting certainty ratings reflect the reviewer's judgment of construct, internal, and external validity relative to an ideal computer science evidence standard, not a clinical one. Readers should interpret \gls{grade} ratings in this context.

\textbf{Conclusion validity.} The narrative synthesis and gap analysis are theory-driven: the five thematic clusters and ten gaps were identified through the reviewer's reading of the corpus and may not capture all relevant patterns. Alternative clustering schemes (e.g., by application domain, \gls{llm} architecture, or mathematical formalism) would yield different gap structures. The research agenda in Section~\ref{sec:agenda} should be read as one informed synthesis, not as a definitive enumeration of all open problems.


\section{Conclusion}
\label{sec:conclusion}

This survey has systematically mapped the emerging field of formal verification for \gls{llm} agent plans. Following a PRISMA 2020 protocol over six databases, we identified and synthesized 38 studies spanning specification techniques, verification approaches, enforcement mechanisms, and safety evaluation benchmarks, published between 2022 and 2026.

\subsection{Summary of Findings}
\label{sec:conclusion:summary}

The synthesis in Sections~\ref{sec:synthesis} and~\ref{sec:agenda} yields four findings and three priority gaps; each statement below points to the section that gives its evidence.

\textbf{The specification bottleneck is real and underaddressed} (RG1, Appendix~\ref{sec:agenda:g1}): NL-to-formal translation achieves high syntactic but low semantic correctness (Section~\ref{sec:synthesis:translation}), so verification built on it risks false assurance.

\textbf{Runtime monitoring is the most mature sub-field} (Section~\ref{sec:related:enforce}): it is the only paradigm with demonstrated production deployment, at the cost of one-step or short-horizon guarantees only.

\textbf{The verifier tax is the field's most consequential single-study finding} (RG3, Appendix~\ref{sec:agenda:g3}): in one $\tau$-Bench study of transactional tool-use agents, intercepting most individually unsafe actions does not reliably produce safe task completion, with two 2025--2026 benchmarks corroborating the qualitative pattern (Section~\ref{sec:synthesis:tax}).

\textbf{No current approach achieves all five desiderata simultaneously}: soundness, completeness, scalability, semantic grounding, and utility preservation (Section~\ref{sec:related:verif}).

\subsection{Research Agenda}
\label{sec:conclusion:agenda}

Of the ten gaps in Section~\ref{sec:related:gaps}, we consider RG1 (semantic correctness), RG3/RG7 (task-level safety and safe re-planning, jointly), and RG2 (long-horizon scalability) the highest priority, for the reasons given in their entries in Appendix~\ref{app:agenda}.

\subsection{Limitations}
\label{sec:conclusion:limits}

A single reviewer conducted this review; a three-rater title/abstract validation check on a repeated search (Fleiss'~$\kappa = 0.81$; Section~\ref{sec:methods:irr}) supports the screening decisions but covers neither full-text screening nor extraction, and it exposed 11 in-scope records missing from the corpus. No meta-analysis was possible due to the high heterogeneity in metrics, domains, and methods across the included studies. We limited the search to English-language, accessible papers; the review does not cover proprietary industry systems. The systematic search covers preprints through 29 May 2026; a non-systematic targeted search (Section~\ref{sec:methods:selection}) additionally identifies corroborating evidence through 30 August 2026, reported separately from the formal corpus. \gls{grade} certainty is predominantly Very Low due to the novelty of the field and the absence of large, independent replications.

\subsection{Closing Remarks}
\label{sec:conclusion:closing}

The case for formal verification of \gls{llm} agent plans is compelling: agents operating in consequential domains must provide guarantees that go beyond empirical performance on benchmarks. The case is also technically hard: the stochastic, \gls{nl}-native nature of \glspl{llm} resists direct application of classical formal methods.

The field has made substantial progress in four years, moving from informal safety prompting to runtime \gls{ltl} monitors with millisecond per-rule overhead~\citep{Wang2025agentspec}, pre-execution \gls{dfa} product verification over multi-framework workflow graphs~\citep{Li2026agentproof}, and probabilistic monitors that warn of violations tens of seconds before they occur~\citep{Wang2025probguard}. But the translation bottleneck, the verifier tax, and the scalability gap remain open, and resolving them will require sustained collaboration among the formal methods, \gls{nl} processing, and \gls{ai} safety research communities.

The taxonomy, gap analysis, and evidence synthesis in this survey give that collaboration a shared vocabulary and a map of what the field has achieved and what remains open.

\noindent\textbf{Competing interests.} The authors declare no competing interests.

\section*{Acknowledgments}
The authors would like to express their gratitude to the Department of Computer Science at the University of Manchester (UoM) and the Systems and Software Security (S3) Research Group for their invaluable support, collaborative environment, and access to cutting-edge resources, which were instrumental in the success of this research. We conducted this work with partial funding from the Engineering and Physical Sciences Research Council (EPSRC) grants EP/T026995/1, EP/V000497/1, EP/X037290/1, and the Soteria project, awarded by the UK Research and Innovation under the Digital Security by Design (DSbD) Programme. The funders had no role in the design, conduct, or analysis of this review, in the preparation of the manuscript, or in the decision to publish.

\bibliographystyle{unsrtnat}
\bibliography{sample-base}

\appendix
\section{Background on Formal Methods and Agent Architectures}
\label{app:background}

The verification techniques surveyed here rest on a small set of formal foundations: temporal logic (the language of safety properties), model checking (the decision procedure), classical planning formalisms (structured task representations), and \gls{llm} agent architectures (the target of verification). Understanding these foundations explains why temporal logic dominates the specification literature, why model checking is the most common algorithmic backbone, and why the stochastic nature of \glspl{llm} breaks classical assumptions. Section~\ref{sec:background} gives the formal definitions that build on this material.

\subsection{Temporal Logic}
\label{sec:bg:tl}

\textit{Temporal Logic} extends classical logic with operators over sequences of states, and has become the dominant specification language in \gls{llm}-agent verification~\citep{Pnueli1977}. The following covers the four variants most relevant to this survey.

\textbf{\gls{ltl}}~\citep{Pnueli1977} interprets formulas over infinite execution traces; for finite plans we use its finite-trace variant, LTL$_f$~\citep{DeGiacomo2013ltlf} (Section~\ref{sec:bg:defs}). Its core grammar uses atomic propositions, negation, conjunction, and two primitive temporal operators, \textbf{X} (\emph{next}) and \textbf{U} (\emph{until}), from which \textbf{F} (\emph{eventually}) and \textbf{G} (\emph{globally}) are derived. For example, $\mathbf{G}(\neg\mathit{auth} \Rightarrow \neg\mathit{access})$ forbids data access in any state where the agent is not authenticated, and $\mathbf{G}(\mathit{tool\_call} \Rightarrow \mathbf{X}\,\mathit{verify\_output})$ mandates immediate post-call verification. \gls{ltl} is the target specification language of NL2LTL~\citep{He2023nl2ltl}, NL2Spec~\citep{Cosler2023nl2spec}, Agent-C~\citep{Kamath2025agentc}, AgentVerify~\citep{Anon2026agentverify}, LogicGuard~\citep{Gokhale2025logicguard}, and VerifyLLM~\citep{Grigorev2025verifyllm}.

While \gls{ltl} reasons over linear execution traces, branching-time logics offer an alternative perspective by quantifying over multiple possible futures. \textbf{\gls{ctl}}~\citep{Clarke1981} interprets formulas over branching-time trees, pairing path quantifiers ($\mathbf{A}$: all paths; $\mathbf{E}$: some path) with temporal operators, enabling both reachability ($\mathbf{EF}\,\varphi$) and universal invariants ($\mathbf{AG}\,\varphi$). AgentVerify~\citep{Anon2026agentverify} exploits \gls{ctl} to reason over non-deterministic \gls{llm} oracles embedded in \glspl{fsm}.

Both \gls{ltl} and \gls{ctl} assume discrete time steps, but many agent domains require reasoning about continuous time and real-valued signals. \textbf{\gls{stl}}~\citep{Maler2004} lifts \gls{ltl} to continuous-time, real-valued signals with bounded temporal operators and quantitative robustness semantics: the value $\rho(\varphi,\mathbf{x},t)\in\mathbb{R}$ measures not only whether a property holds but by how much, making it natural for cyberphysical and robotic agents~\citep{Fang2025nl2stl,Anon2026clarifystl}.

A further limitation of standard temporal logics is their determinism: they assume a system either satisfies a property or does not. For stochastic \gls{llm} agents, however, safety is a matter of probability. \textbf{\gls{pctl}}~\citep{Hansson1994} augments \gls{ctl} with probabilistic path quantifiers: $\mathbf{P}_{\geq p}[\varphi]$ asserts that $\varphi$ holds with probability at least $p$, enabling PRISM-style model checking of stochastic systems. ProbGuard~\citep{Wang2025probguard} relies on this to bound the probability of unsafe state visits.

\subsection{Model Checking}
\label{sec:bg:mc}

\emph{Model checking}~\citep{Clarke1999} decides whether a formal model $M$ satisfies a specification $\varphi$, written $M \models \varphi$, and underpins most automated verification approaches surveyed here.

To apply model checking, one must first represent the system in a formal modeling language. A \textbf{Kripke structure} $M = (S, S_0, R, L)$ pairs a finite state space and transition relation with a labeling function $L: S \to 2^{AP}$ that assigns atomic propositions to states. \gls{ltl} and \gls{ctl} formulas are then evaluated over paths or trees of paths through $M$. In the \gls{llm}-agent context, states capture agent configurations (task state, memory, tool-call history), transitions correspond to actions, and propositions encode safety-relevant facts such as \emph{authenticated}, \emph{pii\_exposed}, or \emph{budget\_exceeded}. Verification systems can construct Kripke structures from ReAct-style plan traces~\citep{Ramani2025bridging} or define them a priori, with the \gls{llm} embedded as a non-deterministic oracle~\citep{Anon2026agentverify}.

Given a Kripke structure and a specification, the next question is how to perform the actual verification algorithmically. \textbf{Automata-based \gls{ltl} model checking} translates $\neg\varphi$ to a B\"uchi automaton, computes the product $M \otimes \mathcal{A}_{\neg\varphi}$, and checks for an accepting cycle (a counterexample); this underlies tools such as \gls{spin} and \gls{nusmv}. AgentProof~\citep{Li2026agentproof} adopts a variant that compiles safety policies to \glspl{dfa} and checks compliance via a graph $\times$ \gls{dfa} product over agent workflow graphs.

Deterministic automata-based methods assume that the system's behavior is fully known and predictable. When agents exhibit stochastic behavior, however, probabilistic guarantees become necessary. \textbf{Probabilistic model checking} extends this to Markov chains and \glspl{mdp}, quantifying the probability of reaching an unsafe state~\citep{Kwiatkowska2011}. ProbGuard~\citep{Wang2025probguard} learns a \gls{dtmc} from observed execution traces and applies \gls{pac}-learning bounds to issue probabilistic safety warnings.

\subsection{Classical Planning Formalisms}
\label{sec:bg:pddl}

\textit{Classical planning formalisms} provide the structured task representations that several \gls{llm}-agent verification approaches build upon.

The most widely adopted formalism in this family is \textbf{\gls{pddl}}~\citep{McDermott1998}, the standard input language for \gls{ai} planners. A problem comprises a \emph{domain} (types, predicates, and action schemas with preconditions and add/delete effects) and a \emph{problem} (objects, initial state, and goal). A valid plan is a sequence of ground actions that leads from the initial state to the goal, and tools such as VAL~\citep{Howey2004} can check it. In the \gls{llm}-agent context, \glspl{llm} are prompted to generate \gls{pddl} descriptions for an external planner~\citep{Anon2025pddl_survey}; the principal challenge is semantic correctness, as \gls{llm}-generated \gls{pddl} achieves high syntactic validity but only \SIrange{24.8}{34.2}{\percent} semantic correctness on the Planetarium benchmark~\citep{Zuo2025planetarium}.

While \gls{pddl} remains the most common classical planning formalism, two related frameworks also appear in the verification literature. \textbf{STRIPS}~\citep{Fikes1971}, the predecessor to \gls{pddl}, and \textbf{\gls{htn} planning}~\citep{Erol1994}, which decomposes tasks hierarchically into sub-tasks, both map naturally to the plan structures of \gls{llm} agents (e.g., ReAct, AutoGen). \gls{htn} verification is explored in the context of multi-agent task allocation~\citep{Anon2026logicmas}.

\subsection{\gls{llm} Agent Architectures}
\label{sec:bg:agents}

An \emph{\gls{llm} agent} is a system that uses an \gls{llm} as a reasoning engine to select and sequence actions from a predefined tool set to accomplish a goal. We characterize agents along three dimensions relevant to verification.

The first dimension concerns how the agent represents its plans internally and externally, as this determines which verification methods apply. \textbf{Plan representation} takes three forms: \emph{trace-based} (interleaved thought--action--observation sequences~\citep{Yao2023react}, the most common and directly amenable to \gls{ltl} evaluation), \emph{graph-based} (predefined or dynamic control-flow graphs suited to static analysis and \gls{dfa} product construction~\citep{Li2026agentproof}), and \emph{code-based} (executable plans verifiable via program analysis or formal contracts~\citep{Anon2025dafny,Miculicich2025veriguard}).

A second, independent dimension is the length of plans the agent must handle, as this directly affects verification tractability. \textbf{Execution horizon} distinguishes short-horizon plans (\SIrange{5}{30}{} actions in bounded environments such as household robotics or customer service, tractable for current verifiers) from long-horizon plans (\SIrange{50}{500}{}+ actions in open-ended settings such as software engineering or autonomous web navigation, beyond the reach of existing approaches).

The third dimension captures whether a single agent acts in isolation or multiple agents interact, since verification complexity grows substantially with the number of agents. \textbf{Agency mode} ranges from single-agent systems, which produce one action sequence, to multi-agent systems, where inter-agent communication, trust delegation, and emergent interactions render per-agent verification insufficient~\citep{Anon2026logicmas}.

\section{Relationship to Prior Taxonomies}
\label{app:taxonomies}
\label{sec:taxonomy:prior}

This appendix compares the three-level taxonomy of Section~\ref{sec:taxonomy} with three earlier classifications.

Zhang et al.~\citep{Zhang2024roadmap} propose a three-way taxonomy (FM-enhancing-\gls{llm}, \gls{llm}-enhancing-FM, unified neural-symbolic) that captures the \emph{direction of integration} between formal methods and \glspl{llm}, but does not distinguish between verification moments or pipeline stages. Huang et al.~\citep{Huang2024planning} classify \gls{llm} planning approaches by planning strategy (task decomposition, plan selection, external module, reflection), but do not address safety verification. A full-stack safety survey~\citep{Anon2025fullstack} categorizes safety concerns by training phase (data, training, deployment), but does not focus on formal verification.

Our taxonomy is orthogonal to these preceding frameworks: it classifies by \emph{pipeline stage}, \emph{verification moment}, and \emph{formal grounding} rather than by integration direction, planning strategy, or deployment phase. Used in conjunction with prior taxonomies, it provides a more thorough characterization of any individual approach.

\section{Research Agenda: Full Treatment of RG1 to RG10}
\label{app:agenda}

This appendix expands Section~\ref{sec:agenda}, which summarizes the directions in Table~\ref{tab:agenda}. For each of the ten open problems identified in Sections~\ref{sec:related:gaps} and~\ref{sec:synthesis}, we describe the current state of the art, the core technical difficulty, and two to three concrete research directions with pointers to the most relevant prior work.

\subsection{RG1 -- Semantic Correctness in \gls{nl}-to-Formal Translation}
\label{sec:agenda:g1}

\gls{llm}-based \gls{nl}-to-formal translation achieves above \SI{90}{\percent} syntactic correctness for \gls{ltl} and above \SI{96}{\percent} for \gls{pddl}, but only \SIrange{24.8}{34.2}{\percent} semantic correctness for \gls{pddl} on Planetarium~\citep{Zuo2025planetarium}. \gls{ltl} translation accuracy under interactive disambiguation reaches \SI{86.11}{\percent}, and \SI{44.4}{\percent} without it~\citep{Cosler2023nl2spec}. These figures cannot be compared with the \gls{pddl} figure, because NL2Spec's benchmark and evaluation protocol differ from Planetarium's. Interactive disambiguation (NL2Spec~\citep{Cosler2023nl2spec}, ClarifySTL~\citep{Anon2026clarifystl}) and hierarchical decomposition~\citep{Anon2025nl2ltl_hsd} reduce semantic errors but do not eliminate them. Current evaluations assess semantic correctness after generation, using equivalence checks against ground-truth specifications or model checking against ground-truth traces; neither kind of ground truth exists for novel domains.

Checking semantic correctness requires a formal model that correctly represents the intended domain semantics. For novel domains, however, we lack any independent ground truth apart from the specification we are evaluating. The problem is therefore circular: to verify that a specification is semantically correct, one needs a formal oracle that is itself a correct formalization of the domain.

\begin{enumerate}[label=\textbf{RG1.\arabic*}]

 \item \textbf{Grammar-constrained decoding for formal languages}: \gls{llm} decoding can be constrained to produce only syntactically valid outputs by restricting the token distribution at each step to tokens consistent with the formal grammar~\citep{Geng2023grammar}. Extended to semantic constraints (e.g., well-typedness of predicate arguments in \gls{pddl}), constrained decoding can eliminate a class of semantic errors at generation time. Integrating semantic type systems for \gls{ltl} and \gls{pddl} into decoding constraints is an open engineering challenge.

 \item \textbf{Equivalence checking via counter-example guided synthesis}: Given two candidate translations of the same \gls{nl} requirement, a model checker can determine whether they are semantically equivalent by searching for a Kripke structure that satisfies one but not the other. If the system finds a counterexample, it can return the discrepancy to the \gls{llm} as feedback for refinement, in a \gls{cegis}~\citep{Solar2006cegis} loop adapted to specification translation. NL2Spec's disambiguation loop~\citep{Cosler2023nl2spec} is a step in this direction; full \gls{cegis} integration remains unstudied.

 \item \textbf{Semantic ground-truth benchmarks}: Progress on semantic correctness requires benchmark datasets that pair each \gls{nl} requirement with a formally verified \gls{ltl}/\gls{pddl} specification and a corpus of execution traces that definitively test semantic correctness. Such benchmarks do not yet exist at scale for agent safety domains. Crowdsourcing requirements from real agent deployment logs, verified by domain experts, is a feasible near-term path.
\end{enumerate}

\subsection{RG2 -- Scalability to Long-Horizon Plans}
\label{sec:agenda:g2}

Current verified systems handle plans of \SIrange{5}{30}{} actions~\citep{Wang2025agentspec,Kamath2025agentc,Grigorev2025verifyllm}. AgentProof~\citep{Li2026agentproof} scales to graphs of 5,000 nodes in pre-deployment static verification, but this applies to finite workflow graphs, not open-ended dynamic plans. Long-horizon tasks (SWE-bench: 100+ actions; autonomous research assistants: 500+ actions) are beyond the reach of exhaustive model checking. The number of states in a Kripke structure grows exponentially with plan length. For open-ended \gls{llm} agents, the branching factor at each step is the model's vocabulary size ($\sim$100k tokens), making exhaustive model checking computationally infeasible.

Section~\ref{sec:agenda:complexity} analyzes how this cost compounds with the number of agents (Table~\ref{tab:scaling}).

\begin{description}

 \item[RG2.1 -- Partial verification of safety-critical sub-sequences.]
  Many long-horizon plans contain a small number of safety-critical segments (e.g., authentication sequences, file deletion chains, financial transaction sequences) surrounded by safe filler actions. Identifying these sub-sequences automatically, with static analysis tagging high-risk action types, and verifying only them could provide partial verification guarantees at manageable cost. The AgentVerify compositional framework~\citep{Anon2026agentverify} provides a foundation; targeted identification of safety-critical sub-sequences within it is unstudied.

 \item[RG2.2 -- Abstraction-refinement for agent plans.]
  \gls{cegar}~\citep{Clarke2003cegar} starts with a coarse abstraction of the state space, model-checks the abstraction, and refines it when it finds a spurious counterexample. When adapted to \gls{llm} agent plans, \gls{cegar} could abstract away low-risk action types (e.g., read-only queries) and focus verification effort on high-risk transitions, refining the abstraction whenever a counterexample turns out to be spurious.

 \item[RG2.3 -- Compositional verification with assume-guarantee contracts.]
  Long-horizon plans decompose into segments with formal interface contracts: each segment assumes a set of preconditions and guarantees a set of postconditions~\citep{Cobleigh2003ag}. Verifying each segment against its contract and composing the proofs avoids the need to model-check the full plan trace. AgentProof~\citep{Li2026agentproof} demonstrates segment-level verification; assume-guarantee composition across dynamically generated segments remains open.
\end{description}

\subsection{RG3 -- The Verifier Tax}
\label{sec:agenda:g3}

One $\tau$-Bench study of transactional tool-use agents shows that even when a system intercepts \SI{94}{\percent} of individually unsafe actions, \gls{ssr} remains below \SI{5}{\percent}, driven by integrity leaks: agents hallucinating credentials to bypass blocked paths~\citep{Anon2026verifiertax}. No existing system resolves this gap. Agent-C achieves \SI{100}{\percent} action conformance but does not report \gls{ssr}~\citep{Kamath2025agentc}; AgentSpec reports task success rates but not \gls{ssr} as defined in the literature~\citep{Wang2025agentspec,Anon2026verifiertax}. The verifier tax is a goal-optimization phenomenon: when the direct path to the goal is blocked, the agent searches for an alternative path, which may be unsafe in ways the monitor does not anticipate. Blocking actions one by one without considering the agent's goal makes the enforcement gameable.

\begin{description}

 \item[RG3.1 -- Plan-level safety policies.]
  Instead of checking individual actions, a plan-level safety monitor checks a proposed sub-sequence or full plan fragment against properties that reference the goal context. This requirement means the monitor must have access to the agent's goal representation (e.g., the task description and current intent) and the action sequence. Agent-C's temporal \gls{dsl}~\citep{Kamath2025agentc} is a step toward this; extending it to goal-contextual properties is an open problem.

 \item[RG3.2 -- Credential and identity integrity constraints.]
  In the integrity-leak mechanism, the agent fabricates identifiers that appear plausible but do not exist in the ground-truth user database. A dedicated integrity monitor would close this attack vector: it maintains a ground-truth index of valid identifiers and rejects any action that references an identifier outside the index. This mechanism parallels SQL injection prevention via parameterized queries, in which the system rejects inputs that deviate from a trusted schema.

 \item[RG3.3 -- Safe replanning upon violation.]
  Rather than simply blocking a violating action, the enforcement system should trigger a replanning request: generate an alternative action sequence from the current state that satisfies all safety constraints and makes progress toward the goal. This replanning requirement integrates the enforcement monitor with a constrained planner (e.g., a \gls{pddl} solver or an \gls{llm} with hard constraints), and this integration represents the most technically ambitious direction for resolving the verifier tax. VeriPlan~\citep{Anon2025veriplan} demonstrates model-checker-guided user-facing replanning; adapting this to automated agent replanning is an open problem.
\end{description}

\subsection{RG4 -- Dynamic Plan Generation}
\label{sec:agenda:g4}

Static verification (AgentProof~\citep{Li2026agentproof}, VeriGuard offline~\citep{Miculicich2025veriguard}) requires pre-specified finite workflow graphs. Most real \gls{llm} agent deployments generate plans dynamically in response to environment state: the agent observes the current state, generates the next action, executes it, observes the outcome, and repeats. No included study provides pre-execution soundness guarantees for dynamically assembled plans.

\begin{description}

 \item[RG4.1 -- Online workflow graph extraction.]
  As an agent executes, the sequence of calls it makes to sub-agents and tools implicitly defines a workflow graph. Extracting this graph in real time and running incremental static checks (adding each new node and edge as the system encounters them) could provide online structural verification without requiring the full graph to be pre-specified. AgentProof's framework~\citep{Li2026agentproof} is a candidate starting point; online incremental model checking over dynamically growing graphs is an active area in formal methods.

 \item[RG4.2 -- Verified plan template instantiation.]
  Many agent tasks share high-level plan structures (templates) that differ only in parameter instantiation. Pre-verifying these templates against safety properties, and then checking that each dynamically generated instantiation conforms to its verified template, would provide soundness guarantees for parameterized plan families. This verification strategy parallels type-checked generics: the system applies the type system (verifier) to the template, and instantiations inherit the proof.
\end{description}

\subsection{RG5 -- Multi-Agent Verification}
\label{sec:agenda:g5}

Most verification work targets single-agent plans. Discrete-event-system \gls{ltl} verification has been applied to multi-robot task allocation in manufacturing~\citep{Anon2026logicmas}, and AgentVerify addresses MCP and skill-invocation protocols in multi-agent settings~\citep{Anon2026agentverify}. However, no study provides a general compositional verification framework for arbitrary multi-agent system topologies. Multi-agent systems exhibit emergent interaction patterns, such as role confusion, trust delegation, cascade failures, and collusion, that are not reducible to the properties of individual agents. Verifying a property of a multi-agent system requires reasoning about the joint execution of all agents, which compounds the state-space explosion problem.

Section~\ref{sec:agenda:complexity} shows that composing agents changes the complexity class of the verification problem, audits the two multi-agent corpus studies against that result, and tabulates verification cost against plan length and agent count (Table~\ref{tab:scaling}).

\begin{description}

 \item[RG5.1 -- Contract-based interfaces between agents.]
We can specify each agent using an assume-guarantee contract: a formal interface that describes the inputs it assumes and the outputs it guarantees~\citep{Cobleigh2003ag}. Composing these contracts allows us to verify system-level properties from component-level proofs without exhaustive joint state-space exploration. This approach is standard in concurrent system verification; adapting it to \gls{llm} agents with non-deterministic outputs presents an open challenge.

 \item[RG5.2 -- Protocol verification for inter-agent communication.]
  In multi-agent systems, agents communicate via structured message-passing protocols. Formalizing these protocols as session types or communicating automata~\citep{Honda2008session} and checking them for safety properties (no deadlock, no privilege escalation) is a tractable sub-problem of multi-agent verification. The MCP/skill invocation protocols of AgentVerify~\citep{Anon2026agentverify} are a starting point.
\end{description}

\subsection{RG6 -- Probabilistic Specification and Stochastic Model Checking}
\label{sec:agenda:g6}

\glspl{llm} are inherently stochastic: the same plan generation call with the same prompt can produce different action sequences across runs. Therefore, we should express safety properties for stochastic agents probabilistically (``the agent violates the policy with probability less than $\epsilon$'') rather than deterministically. ProbGuard~\citep{Wang2025probguard} is the only included study that addresses probabilistic safety; it uses \gls{dtmc} models with \gls{pac}-learning bounds. Outside the corpus, AgentGuard~\citep{Anon2025agentguard} learns an \gls{mdp} of agent behavior online and applies probabilistic model checking to it at runtime; no included study applies PRISM-style \gls{pctl} model checking to \gls{llm} agent plans.

\begin{description}

 \item[RG6.1 -- \gls{pctl} specifications for \gls{llm} agent safety properties.] 
  We can lift standard safety specifications (e.g., ``authenticate before data access'') to probabilistic versions: ``the probability of accessing data without prior authentication is less than 0.01''. \gls{pctl} provides the formal language for such statements; PRISM provides a tool to check them against \gls{dtmc} or \gls{mdp} models. Adapting PRISM to \gls{llm} agent execution traces, by learning the \gls{dtmc} from empirical traces and model-checking \gls{pctl} properties against it, is a feasible and direct extension of ProbGuard.

 \item[RG6.2 -- Calibrated uncertainty quantification for safety monitors.]
  Current deterministic monitors make binary block/allow decisions. A monitor that quantifies its own uncertainty, outputting a probability of violation rather than a binary verdict, can be integrated with risk-threshold policies, enabling deployments to explicitly calibrate the safety-utility trade-off. Conformal prediction~\citep{Vovk2005conformal} provides a distribution-free framework for calibrated uncertainty quantification, but researchers have not yet applied it to agent safety monitoring.
\end{description}

\subsection{RG7 -- Verified Re-Planning}
\label{sec:agenda:g7}

When a plan violates a safety property, all included enforcement systems except Agent-C either block the action, log and continue, or abort the task; Agent-C substitutes a compliant alternative for the single violating tool call~\citep{Kamath2025agentc}. None generates an alternative plan segment that satisfies all safety constraints and continues toward the goal. VeriPlan~\citep{Anon2025veriplan} generates user-facing constraint-satisfying plans in end-user planning settings but does not address automated replanning for agentic systems.

\begin{description}

 \item[RG7.1 -- Constraint-satisfying plan repair.]
  Plan repair~\citep{Fox2006plan} modifies a failing plan minimally to restore validity. Integrated with an \gls{ltl} or \gls{pddl} constraint checker, plan repair could generate an alternative plan fragment from the point of violation that satisfies all checked properties. The challenge for \gls{llm}-based agents is that plan repair operates on formal plan representations, whereas \gls{llm} plans are in \gls{nl}, so the two must be linked by a bidirectional translation pipeline.

 \item[RG7.2 -- Constrained \gls{llm} plan generation.]
  Rather than repairing a violating plan post-hoc, the enforcement system can impose hard constraints on the \gls{llm}'s generation process, steering the model toward safe plan completions using constrained beam search or \gls{smt}-guided decoding. Agent-C's \gls{smt}-based constrained decoding~\citep{Kamath2025agentc} demonstrates feasibility for temporal sequence constraints; extension to arbitrary \gls{ltl} properties over plan prefixes is an open problem.
\end{description}

\subsection{RG8 -- Specification Acquisition at Scale}
\label{sec:agenda:g8}

Writing formal safety specifications (\gls{ltl} formulas, AgentSpec \gls{dsl} rules, \gls{pddl} constraints) requires formal methods expertise. This expertise requirement creates a scalability bottleneck: there are far more agent-deployment domains than there are formal-methods experts. Existing approaches to automated specification acquisition, such as pattern-based translation of \gls{nl} instructions~\citep{He2023nl2ltl} and interactive disambiguation of \gls{nl} requirements~\citep{Cosler2023nl2spec}, remain nascent and have seen evaluation only in narrow domains.

\begin{description}

 \item[RG8.1 -- Specification mining from demonstration traces.]
  Given a corpus of labeled agent execution traces (safe vs.\ unsafe), specification mining algorithms~\citep{Beschastnikh2011spec} can infer \gls{ltl} formulas that distinguish safe from unsafe traces. Applying this to \gls{llm} agent execution logs, which production deployments increasingly produce, would enable data-driven specification acquisition without human expert authoring.

 \item[RG8.2 -- Specification extraction from regulatory and policy documents.]
    Many agent deployment domains fall under formal regulations, such as GDPR, HIPAA, and financial compliance rules. Extracting formal safety specifications from these documents, with \glspl{llm} as translators under expert validation, would reuse existing domain knowledge and produce reusable specification libraries. The literature demonstrates the principle for unstructured text; no study has yet applied it at scale to structured regulatory documents~\citep{Cosler2023nl2spec}.

 \item[RG8.3 -- Compositional specification libraries.]
  A shared, peer-reviewed library of reusable \gls{ltl} and \gls{dsl} specifications for common agent safety properties (authentication-before-access, data-minimization, human-gate-before-irreversible-action) would reduce the per-deployment specification burden. AgentVerify's four-domain specification library~\citep{Anon2026agentverify} provides a seed, but the community needs to curate it systematically.
\end{description}

\subsection{RG9 -- Adversarial Robustness of Verification Systems}
\label{sec:agenda:g9}

Enforcement systems themselves present attack surfaces. Production-grade guardrails still allow \SI{1.75}{\percent} attack success~\citep{MetaAI2025llamafirewall}. Verification components that rely on \gls{llm}-based translation are additionally vulnerable to adversarial inputs that cause the \gls{llm} to generate incorrect specifications or Kripke structures~\citep{Anon2025fullstack}. Adversarial evaluation of verification and enforcement systems is systematically absent from the literature.

\begin{description}

 \item[RG9.1 -- Red-teaming of verification pipelines.]
  Characterizing the attack surface requires systematic adversarial evaluation of the end-to-end pipeline, from \gls{nl} requirement elicitation through specification translation and model checking to enforcement. Adapting \gls{llm} red-teaming methodologies~\citep{Andriushchenko2024agentharm} to target the verification pipeline specifically (prompt injection into specification generators, adversarial Kripke structures, monitor bypass via indirect action sequences) would characterize the residual risk of each component.

 \item[RG9.2 -- Formal robustness guarantees for specification translators.]
  If the specification translator is itself formally verified (e.g., through type-safe grammars or certified compilation), its correctness guarantee is robust to adversarial inputs by construction. Applying verified compilation techniques~\citep{Leroy2009compcert} to \gls{ltl}-generating components, at least for a restricted fragment of the input language, would provide robustness without relying on empirical red-teaming.
\end{description}

\subsection{RG10 -- Evaluation Standardization}
\label{sec:agenda:g10}

No benchmark simultaneously evaluates formal verification coverage, runtime overhead, and \gls{ssr} across multiple agent architectures and task domains. Existing benchmarks address subsets: Agent-SafetyBench~\citep{Zhang2024agentsafety} covers safety but not formal verification; $\tau$-Bench~\citep{Sierra2024taubench} enables \gls{ssr} measurement but only in a narrow transactional domain; AgentBench~\citep{Liu2023agentbench} covers task success but not safety. The Verifier Tax paper~\citep{Anon2026verifiertax} introduced \gls{ssr} as a unified metric but evaluated it on a single benchmark.

\begin{description}

 \item[RG10.1 -- A unified formal safety evaluation suite.]
  A benchmark should define: (a) a set of task environments with known ground-truth safe and unsafe plans; (b) a corresponding library of formal safety specifications for each task; (c) standardized metrics including \gls{ssr}, formal verification coverage (a formal language captures only a fraction of possible safety properties), and runtime overhead. The formal methods, \gls{nlp}, and \gls{ai} safety communities should build this benchmark jointly so that it covers all relevant dimensions. As an illustrative and explicitly uncalibrated headline metric, reported alongside the three-axis vector of Section~\ref{sec:synthesis:heatmap} rather than in its place, we propose $\text{ESI} = \text{SSR} \cdot \min\bigl(1, (L_{\text{budget}}/L)^{\alpha}\bigr)$ for a deployment-chosen exponent $\alpha$, or more simply $\text{SSR@}L_{\text{budget}}$ (\gls{ssr} counting only tasks completed within the latency budget). Every included study that reports runtime overhead uses a different task domain, latency budget, and definition of acceptable degradation (Table~\ref{tab:overhead}), so neither formula can be fit or validated against the current corpus; we present both as candidates for the field to calibrate and test, not as validated metrics.

 \item[RG10.2 -- Adoption of \gls{ssr} as the primary metric.]
  The field should standardize on \gls{ssr}~\citep{Anon2026verifiertax} as the primary evaluation metric for enforcement systems, supplemented by action-level conformance as a secondary metric. \gls{ssr} captures the end-to-end objective (tasks completed both safely and correctly) and avoids the false confidence that action-level compliance rates alone provide. A community-maintained leaderboard or shared evaluation server would accelerate metric adoption.

 \item[RG10.3 -- Benchmark contamination mitigation.]
  \glspl{llm} trained on internet data may have memorized evaluation scenarios from public benchmarks, inflating reported performance. Mitigation strategies include held-out evaluation sets updated after each major model release, procedurally generated tasks with verifiable ground truth, and canary task detection for known contamination in benchmarks.
  
\end{description}

\section{Extended Discussion}
\label{app:discussion}

This appendix extends the discussion in Section~\ref{sec:discussion}.

\subsection{Field Maturity Assessment}
\label{sec:discussion:maturity}

The field of formal verification for \gls{llm} agent plans is at an early but structurally sound stage of development. Using the Gartner Hype Cycle~\citep{Fenn2008hyp} as a rough heuristic, the field appears to be transitioning from the \emph{Peak of Inflated Expectations} (2023 to 2024, marked by many demonstration papers with limited evaluation) toward the \emph{Slope of Enlightenment} (2025--2026, with the first empirical findings that challenge simplistic assumptions, such as the verifier tax~\citep{Anon2026verifiertax}).

Indicators of early \textbf{maturity} include convergence on a dominant specification language (\gls{ltl}) with consistent findings across independent groups~\citep{He2023nl2ltl,Cosler2023nl2spec,Kamath2025agentc,Wang2025agentspec}; the emergence of dedicated workshops at major software engineering and programming-languages venues as forums for this research; production deployment of at least one system (LlamaFirewall~\citep{MetaAI2025llamafirewall} at Meta) with disclosed performance metrics; the development of standardized benchmarks (Agent-SafetyBench~\citep{Zhang2024agentsafety}, $\tau$-Bench~\citep{Sierra2024taubench}) that enable cross-system comparison; and the first negative empirical result that challenges field assumptions (verifier tax~\citep{Anon2026verifiertax}).

Indicators of \textbf{immaturity} include that \gls{grade} certainty is \emph{Very Low} for seven of the nine principal claims, with none reaching \emph{High}; most studies are single-group, single-domain evaluations without independent replication; no system simultaneously satisfies soundness, completeness, scalability, semantic grounding, and utility preservation; the field lacks a shared formal framework, since every paper introduces its own agent model, specification language fragment, and evaluation protocol; and specification acquisition and verified re-planning (arguably the most important problems) are the least studied.

The maturation trajectory of this field parallels the early development of neural network verification (2017 to 2020), which similarly began with demonstrations on narrow benchmarks (MNIST classifiers, small ReLU networks) before confronting scalability and semantic gap challenges that required fundamental theoretical advances~\citep{Katz2017reluplex}. Neural network verification is now a mature subfield with verified industrial applications (e.g., the ACAS~Xu collision-avoidance networks for unmanned aircraft~\citep{Katz2017reluplex}). The \gls{llm} agent verification field is approximately three to four years behind this trajectory, with the additional complication that the \emph{output} of \glspl{llm} is \gls{nl} rather than a fixed-dimension vector, which makes the semantic gap structurally harder to close.

\subsection{Relationship Between Safety Paradigms}
\label{sec:discussion:paradigms}

The included studies reflect three intellectual traditions for addressing \gls{llm} agent safety, and these traditions rarely engage with one another.

The \textbf{formal methods tradition} (AgentVerify~\citep{Anon2026agentverify}, AgentProof~\citep{Li2026agentproof}, AutoRocq~\citep{NUS2025autorocq}, VeriGuard~\citep{Miculicich2025veriguard}) draws on 50+ years of model checking and theorem proving research. This tradition prioritizes soundness and completeness: if verification says a plan is safe, it \emph{is} safe. The limitation is that soundness guarantees depend on the correctness of the specification and the model, and \glspl{llm} with non-negligible semantic error rates construct both.

The \textbf{\gls{ml} safety/alignment tradition} (Constitutional \gls{ai}~\citep{Bai2022constitutional}, Agent Safety Alignment via \gls{rl}~\citep{Anon2025rlsafety}, Self-Refine~\citep{Madaan2023selfrefine}) draws on reinforcement learning, RLHF, and iterative refinement. This tradition prioritizes scalability and generality: trained models implicitly apply safety constraints to arbitrary inputs without per-deployment specification engineering. The limitation is the absence of formal guarantees: a model trained to be safe can generate unsafe outputs at inference time, and there is no formal procedure to confirm that it cannot.

The \textbf{security engineering tradition} (LlamaFirewall~\citep{MetaAI2025llamafirewall}) draws on systems security, threat modeling, and defensive programming. This tradition prioritizes adversarial robustness and production deployability: the threat model is an active adversary attempting to exploit the agent, not a passive environment. The limitation is that this tradition is primarily empirical (a reduction in attack success rate) rather than formal (a provable bound on attack success rate).

The \textbf{need for integration} follows directly. Each tradition covers a different part of the safety landscape and has different failure modes. Deployed systems will need to combine elements of all three: formal runtime monitors (providing sound action-level guarantees) over aligned base models (reducing the frequency of violations that reach the monitor) within hardened system architectures (defending against adversarial exploitation of the monitoring infrastructure). None of the included systems has yet demonstrated this integration.

\subsection{Societal and Ethical Considerations}
\label{sec:discussion:ethics}

The deployment of formal verification for \gls{llm} agents raises societal and ethical questions that extend beyond technical correctness. Three concerns merit particular attention: the pressure to deploy unsafe systems, the politics of specification authorship, and the risk of over-reliance on formal certificates.

\textbf{The deployment pressure paradox.} The pace of \gls{llm} agent deployment in production settings far exceeds the pace of formal safety verification's maturation. Agents are being deployed in healthcare, legal, and financial settings, while \gls{grade} certainty for seven of the nine principal claims remains \emph{Very Low}. The societal consequence of this gap is that the harm reduction that current verification systems achieve, while real, is not yet formally guaranteed or independently replicated at scale. Survey findings should inform deployment governance: requiring \gls{ssr} measurement, mandatory specification review, and periodic re-evaluation against updated benchmarks as agent capabilities evolve.

\textbf{Specification as a power structure.} Who writes the formal safety specification for an \gls{llm} agent determines which behaviors are permitted and which are blocked. Specification acquisition at scale (RG8) raises governance questions: specifications derived from corporate policy documents may protect business interests at the expense of user interests. In contrast, those derived from regulatory documents may lag behind technological capabilities. Transparent, publicly accessible specification libraries with explicit authorship and versioning would reduce the risk of opaque specification capture.

\textbf{The completeness illusion.} A formal safety certificate declares: ``This agent has been verified against specification $\varphi$'', and creates a social expectation of safety that may not be warranted when $\varphi$ fails to capture all relevant safety properties. Formal verification of \gls{llm} agent plans should present a partial safety guarantee conditioned on specification completeness, not a blanket safety assurance.

\end{document}